\documentclass[12pt,journal,letterpaper,onecolumn]{IEEEtran}

\ifCLASSINFOpdf
\else
\fi
\usepackage{amssymb}
\usepackage{amsmath}
\usepackage{amsfonts}
\usepackage{graphicx}
\usepackage{amsthm}
\usepackage{algorithm}
\usepackage{algorithmic}
\usepackage{mathtools}
\usepackage{lettrine}
\usepackage{subfigure}
\usepackage{cite}

\DeclarePairedDelimiter{\ceil}{\lceil}{\rceil}
\newcommand{\quotes}[1]{``#1''}

\newtheorem*{theorem*}{Theorem}

\begin{document}
%
\title{Multiple Instance Fuzzy Inference Neural Networks}
%
%
%

\author{Amine~B.~Khalifa\footnote{Email: a0benk01@louisville.edu\\“This work has been submitted to the IEEE for possible publication. Copyright may be transferred without notice, after which this version may no longer be accessible.”}, Hichem~Frigui,~\IEEEmembership{Member,~IEEE},\\Multimedia Research Lab\\
CECS Department\\University of Louisville\\ Louisville, KY 40292, USA}

\maketitle
\begin{abstract}
Fuzzy logic is a powerful tool to model knowledge uncertainty, measurements imprecision, and vagueness. However, there is another type of vagueness that arises when data have multiple forms of expression that fuzzy logic does not address quite well. This is the case for multiple instance learning problems (MIL). In MIL, an object is represented by a collection of instances, called a bag. A bag is labeled negative if all of its instances are negative, and positive if at least one of its instances is positive. Positive bags encode ambiguity since the instances themselves are not labeled. In this paper, we introduce fuzzy inference systems and neural networks designed to handle bags of instances as input and capable of learning from ambiguously labeled data. First, we introduce the Multiple Instance Sugeno style fuzzy inference (MI-Sugeno) that extends the standard Sugeno style inference to handle reasoning with multiple instances. Second, we use MI-Sugeno to define and develop Multiple Instance Adaptive Neuro Fuzzy Inference System (MI-ANFIS). We expand the architecture of the standard ANFIS to allow reasoning with bags and derive a learning algorithm using backpropagation to identify the premise and consequent parameters of the network. The proposed inference system is tested and validated using synthetic and benchmark datasets suitable for MIL problems. We also apply the proposed MI-ANFIS to fuse the output of multiple discrimination algorithms for the purpose of landmine detection using Ground Penetrating Radar.

%
%
\end{abstract}


%
\IEEEpeerreviewmaketitle

\section{Introduction}
Fuzzy inference is a powerful modeling framework that can handle computing with knowledge uncertainty and measurements imprecision effectively \cite{Zadeh1977}. It 
 has been successfully applied to a wide range of problems, mainly in system modeling and control \cite{Zadeh1973, ChenDynamicSystems,Jager171945}. Most of the proposed fuzzy inference methods gained success because of their ability to leverage expert knowledge to identify the model parameters \cite{Lee52552}. This practice simplifies system design and ensures that the knowledge base (if-then rules) used by the system is easy to interpret \cite{casillas2003interpretability}.

More recently, fuzzy inference has increasingly been applied to more advanced applications, such as content-based information retrieval \cite{CBIR}, image segmentation \cite{Othman6461091}, image annotation \cite{Imageannotation5617668}, pattern recognition \cite{pt952888}, recommender systems \cite{Adnan6850800}, and multiple classifier fusion \cite{KhalifaSPIE2014}. The aforementioned applications are more challenging as they require extensive knowledge base to accommodate for various scenarios. Since this diverse knowledge base cannot be fully captured by domain experts, data-driven techniques are typically used to identify and learn the inference system's parameters \cite{Rulebase1007339,Datadriven1630151}. One such technique is the Adaptive Neuro-Fuzzy Inference System (ANFIS)\cite{Anfis}. ANFIS is a universal approximator that combines the learning and modeling power of neural networks and fuzzy logic into an adaptive inference system. It is a hybrid intelligent system and it provides a systematic approach to jointly learn the optimal input space partition (rules) and the optimal output parameters using supervised learning.

Typically, in supervised learning, access to large labeled training datasets improves the performance of the devised algorithms by increasing their robustness and generalization capabilities. Nowadays, access to such large datasets is becoming more convenient. However, for a supervised leaning method to benefit from this data, it need to be carefully preprocessed, filtered, and labeled. Unfortunately, this process can be too tedious as the vast portion of the collected data is unstructured, labeled ambiguously and at a coarse level. An alternative and a relatively new framework of learning that tackles the inherent ambiguity better than supervised learning, is the  Multiple Instance Learning (MIL) paradigm \cite{maron1998learning}.
\subsection{Multiple Instance Learning}\label{MILintro}
Unlike standard supervised learning, in MIL, an object is not represented by a simple data point, but rather by a collection of instances, called a bag. Each bag can contain a different number of instances. A bag is labeled negative if all of its instances are negative, and positive if at least one of its instances is positive\footnote{Note that
positive bags may also contain negative instances.}. Positive bags can encode ambiguity since the instances themselves are not labeled. Given a training set of labeled bags, the goal of MIL is to learn a concept that predicts the labels of training data at the instance level and generalizes to predict the labels of testing bags  and their instances\cite{Dietterich1997}. We refer to this definition as the standard MIL assumption. Multiple MIL paradigms have been proposed \cite{alpaydin2015single}, but for simplicity we focus our formulation on the standard MIL assumption. 

The MIL is a well known problem that has been studied for the last 20 years, it was first formalized by Dietterich et al. \cite{dietterich1997solving} providing a solution to drug activity prediction. Ever since, it has increasingly been applied to a wide variety of tasks including content-based information retrieval \cite{Retrieval4475952}, drug discovery \cite{Maron}, pattern recognition \cite{karem2011multiple}, image classification \cite{rahmani2006missl}, region-based image categorization \cite{MILES1717454}, image annotation \cite{yang2005region}, object tracking \cite{babenko2011robust} and time series prediction \cite{maron1998learning}. In general, MIL can be applied in two contexts of ambiguity: \quotes{polymorphism ambiguity} and \quotes{part-whole ambiguity} \cite{Andrews}. In polymorphism ambiguity, an object can have multiple forms of expression in the input space and it is not known which form is responsible for the object label. Whereas, in part-whole ambiguity, an object can be broken into several parts represented by different feature vectors in the input space. However, only few parts are responsible for the object label \cite{babenko2008multiple}. Polymorphism Ambiguity arise more often in applications related to chemistry and bioscience. The original MIL application of drug discovery \cite{Dietterich1997,maron1998learning} is a case of polymorphism ambiguity. Part-whole Ambiguity is more common in pattern recognition problems. For example, in image annotation features are usually extracted locally (from patches) while the labels, or tags, are only available gloablly at the image level. Another closely related application is object detection. In this application, objects of interest may cover only a limited region of the image, the rest could be other objects or background. Traditional supervised learning requires identifying image patches containing the object of interest only and labeling them. As indicated by Viola et al. \cite{zhang2005multiple}, placing bounding boxes around objects is an inherently ambiguous task. Thus, to avoid the tedious task of object segmentation and annotation, the problem of object detection can be addressed using an MIL paradigm. To illustrate the need for MIL further, in the following we analyze how a multiple instance (MI) representation can be applied to image classification. More details about MIL taxonomy have been reported by Amores \cite{amores2013multiple}.

\begin{figure}[htb]
  \centering
  \includegraphics[width=1\linewidth]{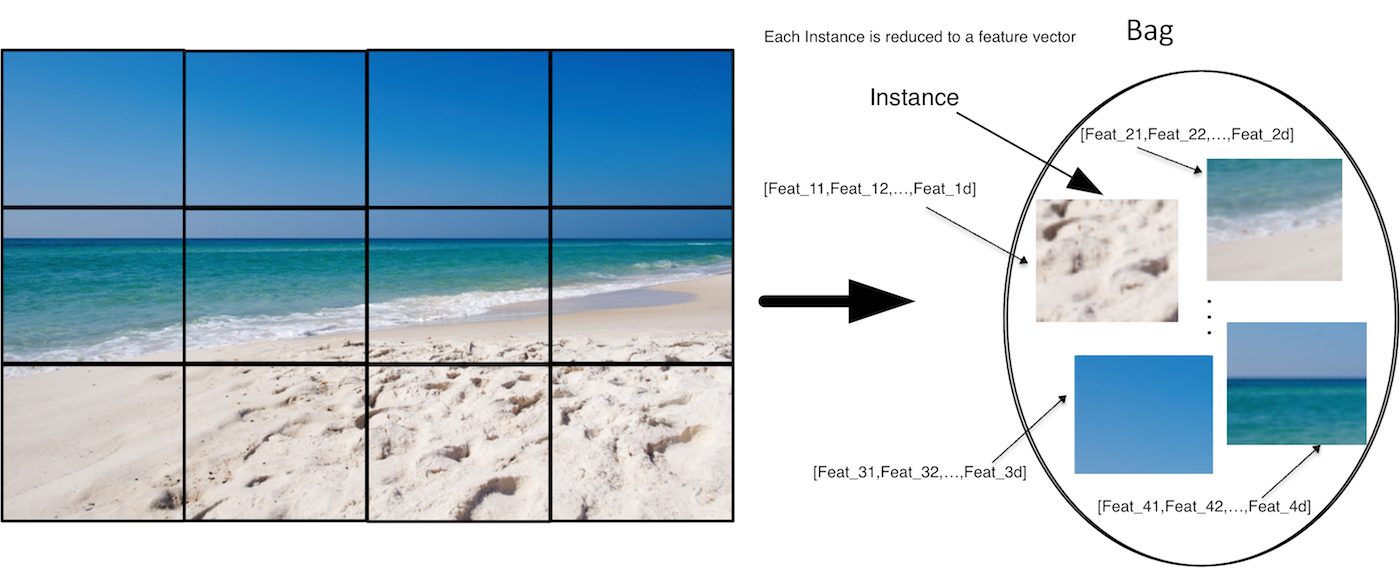}
   \caption{Example of an image represented as a bag of 12 instances. Each instance correspond to a feature vector (e.g., color, texture) extracted from one patch. The bag is labeled ``sky'' because at least one of its instances is sky. However, many other instances are not ``sky''. Labels at the instance level are not available.}
  \label{fig:BAG-Example}
\end{figure}

Consider the simple example of classifying images that contain \quotes{sky}. Using an MIL approach, each training image is represented by a bag of instances where each instance corresponds to features extracted from a region of interest. These regions could be obtained by segmenting the image or simply by dividing it into patches. A multiple instance representation is well suited for this purpose because only few regions may contain the object of interest (sky), that is the positive class. Other patches will be from background or other classes. This representation is illustrated in Figure \ref{fig:BAG-Example}. Traditional single instance learning are based on instance level (patch-level) labels and would require each image region to be correctly segmented and labeled prior to learning.
\subsection{Fuzzy Inference Systems}
A Fuzzy Inference System (FIS) is a paradigm in soft computing which provides a means of approximate reasoning \cite{lanzi2000learning}. A FIS is capable of handling computing with knowledge uncertainty and measurements imprecision effectively \cite{Zadeh1977}. It performs a non-linear mapping from an input space to an output space by deriving conclusions from a set of fuzzy if-then rules and known facts \cite{Cordón2011894}. Fuzzy rules are condition/action (if-then) rules composed of a set of linguistic variables (e.g. image patch). Each variable is assigned a linguistic term (e.g. red, green, blue). For instance, the following rules could be used to identify patches from the image in Figure \ref{fig:BAG-Example}:
\begin{itemize}
  \item If \textit{patch} is \textit{blue} and \textit{texture} is \textit{smooth} then \textit{region} is \textit{sky}.
  \item If \textit{patch} is \textit{blue} and \textit{patch position} is \textit{upper half}  then \textit{region} is \textit{sky}.
\end{itemize}

Typically, a FIS is composed of 5 components: (1) a \textit{Fuzzification} unit that assigns a membership degree to each crisp input dimension in the input fuzzy sets; (2) a \textit{Knowledge Base} characterized by fuzzy sets of linguistic terms; (3) a \textit{Rule Base} containing a set of fuzzy if-then rules; (4) an \textit{Inference unit} that performs fuzzy reasoning; and (5) a \textit{Deffuzification unit} that generates crisp output values.
FIS has proven to be very effective in various applications \cite{Zadeh1973,Mamdanicontrol, Babuška19961593, ChenDynamicSystems,Mizumoto1988129,Lee52551,Sugeno390281,Yager247902,Tacker4045839,Singh6659349,Jager171945}. However, it is not applicable to cases where objects are represented by multiple instances.


\subsection{Motivations For Multiple Instance Fuzzy Inference}
There are two major limitations that prevent using standard FIS methods with multiple instance data. First, due to the absence of labels at the instance level, we cannot use standard FIS learning methods to construct the knowledge base. Second, we need an effective mechanism to aggregate instances' confidences and infer at the bag level. The above limitations are due mainly to the inherent architecture of fuzzy inference systems.  The standard inference systems reason with individual instances. First, the system's input is an individual instance. Second, the rules describe fuzzy regions within the instances space. Third, the output of the system corresponds to the fuzzy inference using  a single instance. Fourth, labels of the individual instances are required when using learning techniques to identify the parameters of the system. In summary, traditional fuzzy inference systems cannot be used effectively within the MIL framework.

To address the above limitations, we introduce two FIS designed to handle reasoning with bags of instances and capable of learning form ambiguously labeled data. The first one, called Multiple Instance-Sugeno (MI-Sugeno) extends the standard Sugeno system \cite{Sugeno}. The second one, called Multiple Instance-ANFIS (MI-ANFIS) extends the standard ANFIS \cite{Anfis} system and uses MI-Sugeno rules. We report results on various experiments and discuss the advantages of using our proposed methods over closely related MIL algorithms such as Multiple Instance Neural Networks \cite{zhou2002neural} (MI-NN) and Multiple Instance RBF Neural Networks \cite{zhang2006adapting} (RBF-MIP).

\section{Multiple Instance Fuzzy Inference}\label{MI-fuzzInference}
In the following, let $B_p$ be a bag of $M_p$ instances with the $j$th instance denoted as $\mathbf{x}_{pj} \in \mathbb{R}^D$ with elements $x_{(p,j,k)}$ corresponding to features, i.e.,
\begin{equation}\label{Bag}
    B_{p}=\left(
      \begin{array}{c}
        \mathbf{x}_{p1} \\
        \mathbf{x}_{p2}  \\
        \vdots \\
        \mathbf{x}_{pM_p} \\
      \end{array}
    \right)=\left(
      \begin{array}{cccc}
        x_{(p,1,1)} & x_{(p,1,2)} & \ldots & x_{(p,1,D)}\\
        x_{(p,2,1)} & x_{(p,2,2)} & \ldots & x_{(p,2,D)} \\
        \vdots & \vdots & \ddots & \vdots \\
        x_{(p,M_p,1)} & x_{(p,M_p,2)} & \ldots & x_{(p,M_p,D)} \\
      \end{array}
    \right).
\end{equation}
Note that the number of instances can vary between bags ($M_p$ depends on $B_{p}$). A bag is labeled positive if at least one of its instances is positive, and negative if all of its instances are negative.\\
\subsection{Multiple Instance Sugeno Style Fuzzy Inference}\label{MI-Sugeno}

To adapt Sugeno inference to problems where objects are described by multiple instances, we propose a multiple instance Sugeno inference (MI-Sugeno) system that uses multiple instance fuzzy if-then rules.
Recall that a fuzzy if-then rule is expressed as
\begin{equation}\label{eq:tradional_if-then_rule}
    if\;x\;is\;A\;then\;y\;is\; C
\end{equation}
where $A$ and $C$ are fuzzy sets on universes of discourse $X$ and $Y$, respectively.
The rule in (\ref{eq:tradional_if-then_rule}) combines the fuzzy propositions ($x\;is\;A$, $y\;is\;C$) into a logical implication abbreviated as $A\rightarrow C$ with membership function $\mu_{A\rightarrow C}(x,y)$.
The rule is defined using a premise part that is a single instance fuzzy proposition.
\\To generalize the rule in (\ref{eq:tradional_if-then_rule}) to MI data, we define a multiple instance fuzzy rule as:
\begin{equation}\label{eq:fuzzy_if-then_rule}
    if\;B_i\;is\;A\;then\;y\;is\;C  \Longleftrightarrow if\;\bigvee_{j=1}^{M_i}(\mathbf{x}_{ij}\;is\;A)\;then\;y\;is\;C
\end{equation}
where as in (\ref{eq:tradional_if-then_rule}), $A$ and $C$ are fuzzy sets on the universes of discourse $X$ and $Y$, respectively. In (\ref{eq:fuzzy_if-then_rule}), $B_i$ is a bag of instances $\mathbf{x}_{ij}$ as defined in (\ref{Bag}), and $M_i$ is the number of instances in $B_i$. The premise part of a multiple instance fuzzy rule (i.e., $\bigvee_{j=1}^{M_i}(\mathbf{x}_{ij}\;is\;A)$ ) is a multiple instance proposition, whereas the consequent part is a traditional proposition. In (\ref{eq:fuzzy_if-then_rule}), $\bigvee$ is a joint operator that can be any T-conorm (maximum, algebraic sum, bounded sum, etc.). The reason behind using a T-conorm for combining individual instances' responses, goes back to the standard MIL assumption \cite{maron1998learning,Dietterich1997} which states that a bag is positive if and only if one or more of its instances are positive. Thus, the bag-level class label is determined by the disjunction of the instance-level class labels. We note that the T-conorm can be designed to handle a broader set of non-standrad MIL problems, for example to allow the inference process to assign a higher degree of belief to bags with more than one positive instance.

The proposed MI-Sugeno uses multiple instance fuzzy rules with a consequent part that is described by means of a function $C$ that maps a bag of instances to a crisp numerical value. Specifically, we define a multiple instance sugeno rule as:
\begin{equation}
\begin{split}
\mathcal{R}^i(\mathbf{B}_p):\bigvee_{j=1}^{M_p}(  \; \mbox{If} \; x_{(p,j,1)} \; \mbox{is} \; A^i_1 \; \mbox{and} \; x_{(p,j,2)}  \; \mbox{is} \; A^i_2\ldots\mbox{and} \;x_{(p,j,D)} \; \mbox{is} \; A^i_D), \; \mbox{then} \; \\o^i=C(\mathbf{x}_{p1}\cdot \mathbf{b}^i,\;  \mathbf{x}_{p2}\cdot \mathbf{b}^i, \ldots, \mathbf{x}_{pM_p}\cdot \mathbf{b}^i)
\label{eq:mirule}
\end{split}
\end{equation}
\begin{figure}[!t]
  \centering
  \includegraphics[width=1\linewidth]{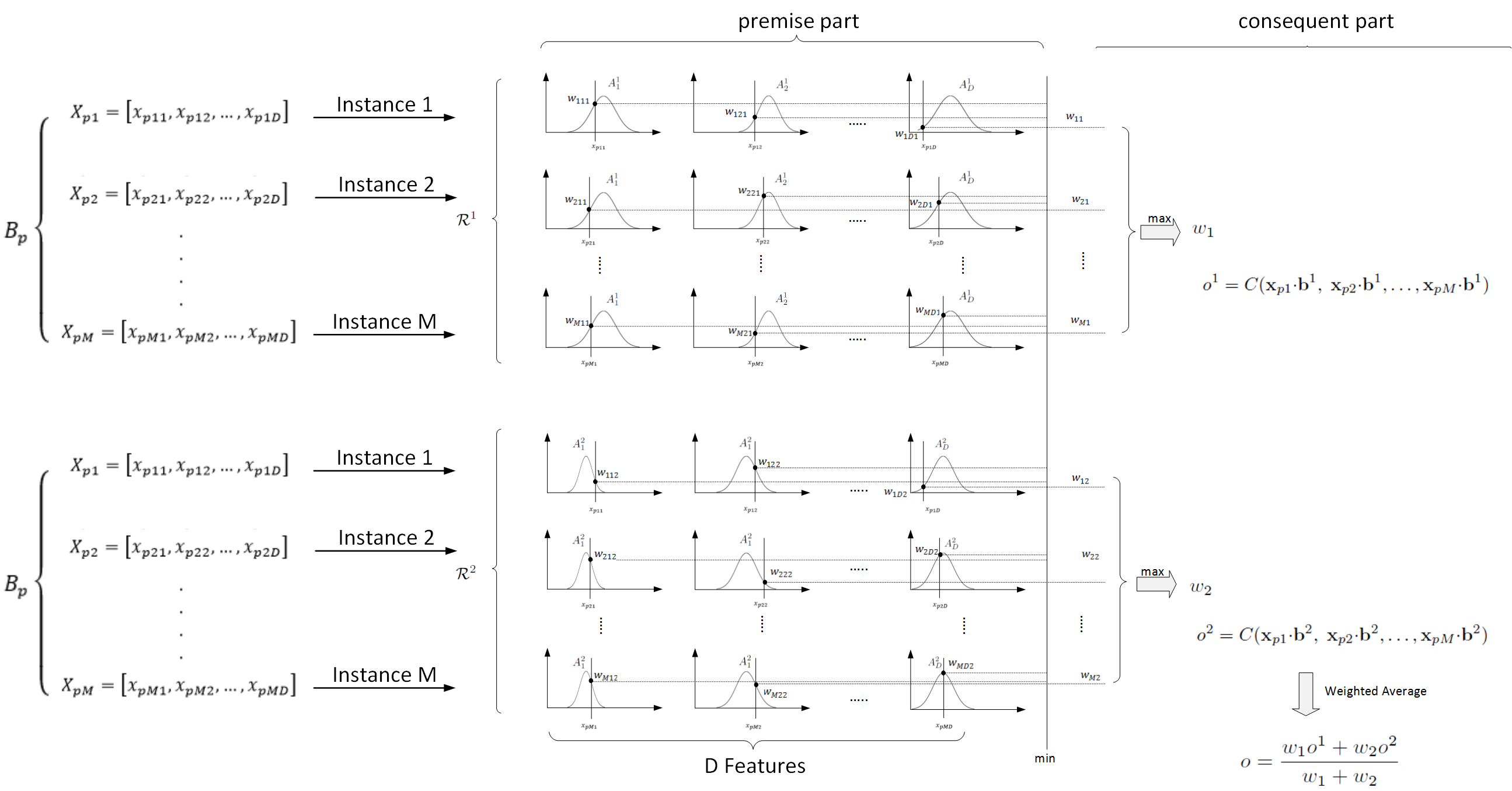}
   \caption{Illustration of the proposed multiple instance Sugeno fuzzy inference system with 2 rules.}
  \label{fig:MI-Sugeno}
\end{figure}
In (\ref{eq:mirule}), $\mathbf{b}^i=b_0^i,...,b_D^i$ is a set of polynomial coefficients. When the polynomial coefficients $\mathbf{b^i}$ are first order, the MI-Sugeno fuzzy model is called \textbf{first order}, and \textbf{zero order} when the polynomial coefficients are zero order.\\
Figure \ref{fig:MI-Sugeno} illustrates the proposed MI-Sugeno system and its fuzzy inference mechanism to derive the output, o, in response to a bag of $M$ instances for the simple case of two rules. The premise part of the rules evaluates all the bag's instances simultaneously. The inference starts by the fuzzification of instances $\mathbf{x}_{pm}$ of input bag $B_p$. Fuzzification assigns a membership degree to each input instance dimension in the rules input fuzzy sets. In Figure \ref{fig:MI-Sugeno}, instance $\mathbf{x}_{pm}$ activates the $i$th input fuzzy set of the $j$th rule by a degree of truth $w_{(m,i,j)}$. Next, an implication process is executed to combine the activations of the instances within the bag resulting in the activation of the rules' output with different degrees. In this example, we use a simple min operator, and the output of rule $R^j$ will be partially activated by a degree $w_{mj}$ = $min_{k=1,\dots,D}w_{(m,k,j)}$. The $w_{mj}$ (truth instances) are combined in the premise part using the max T-conorm, resulting in the activation of rule $R^j$ by a degree $w_{j}$ = $max_{m=1,\dots,M}\{w_{mj}\}$. To evaluate the consequent part, first the linear response of each instance is computed, i.e., $\mathbf{x}_{pj}\cdot \mathbf{b}^i$. Then, a function $C$ is used to compute the final output by combining the instances' responses. Many functions could be used and the choice should be domain-specfic. The output of each rule, $o^1$ and $o^2$, are crisp values. As in the traditional Sugeno fuzzy inference system, the overall output of the system is obtained by taking the weighted average of the rules' outputs.

The consequent part of the proposed MI-Sugeno style inference system is inspired by the work of Ray and Page on multiple instance regression \cite{ray2001multiple}. In their work, the authors proposed a regression framework for predicting bags' labels. This formulation allows the linear coefficients $\mathbf{b}^i$ and the parameters of the combining function $C$ to be learned using optimazation techniques, as we will show in section \ref{MI-ANFIS}.
\\Similar to traditional fuzzy inference, the premise part of a multiple instance rule defines a local fuzzy region within the instance space, and the consequent part describes the characteristics of the system's output within each region. More specifically, in problems, a local region describes a positive concept (also called target concept), and the output of a rule represents the degree of \quotes{positivity} of the instances in that target concept. A target concept is a region in the instances' feature space that includes as many instances from positive bags as possible and as few instances from negative bags as possible.

The Sugeno fuzzy model \cite{Sugeno} was the first attempt at learning fuzzy rules from training data. It has been used to develop the standard ANIFS which combines the representation power of fuzzy inference and learning capability of neural networks to learn the rules. In the next section, we will use our MI-Sugeno to develop a multiple instance extension of ANFIS (MI-ANFIS). 

\subsection{MI-ANFIS: A Multiple Instance Adaptive Neuro-Fuzzy Inference System}\label{MI-ANFIS}

\begin{figure}[!t]

  \centering
  \includegraphics[width=1\linewidth]{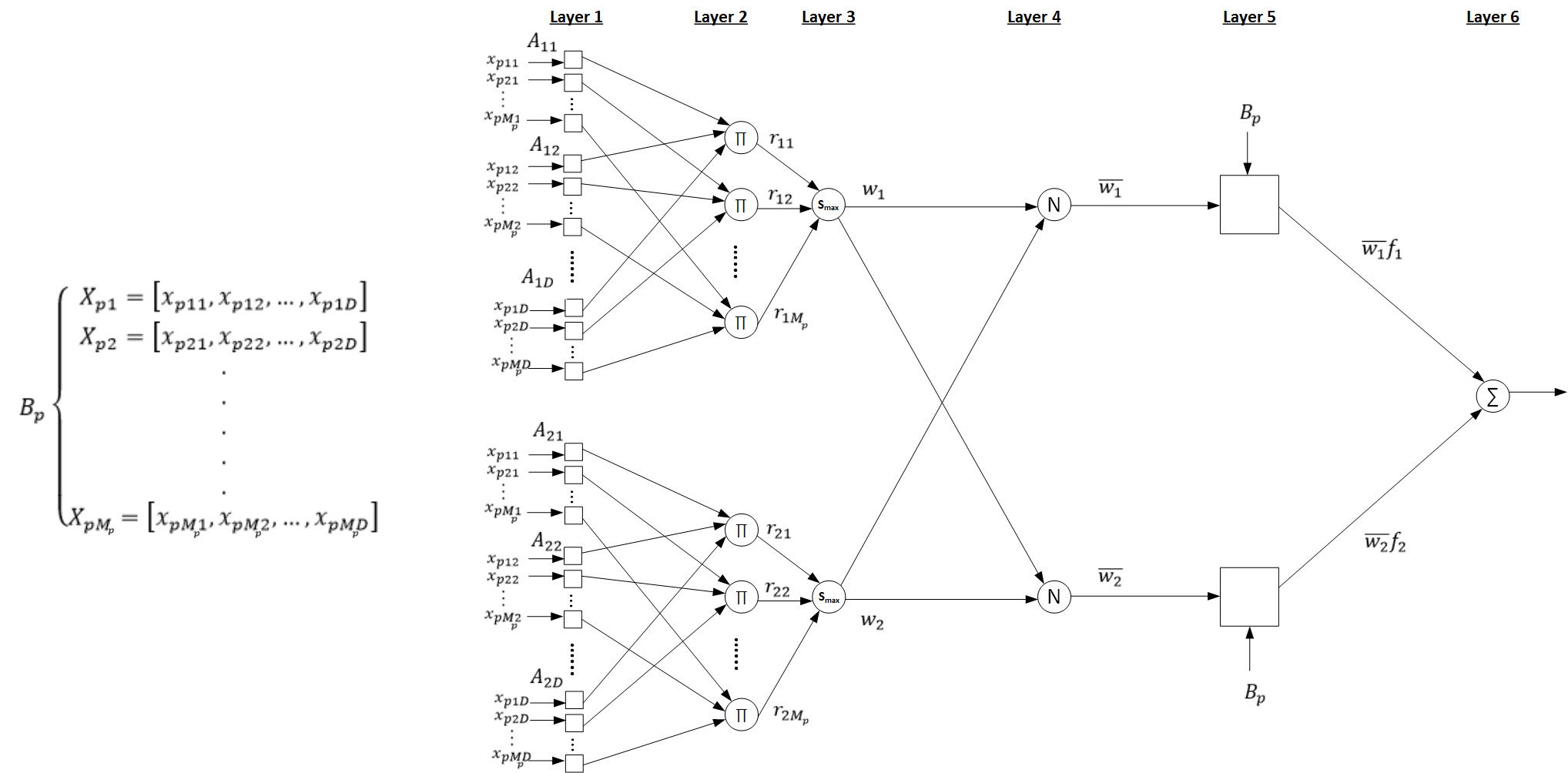}
   \caption{Architecture of the proposed Multiple Instance Adaptive Neuro-Fuzzy Inference System}
  \label{fig:MI-Anfis-arch}

\end{figure}

Let $B_i$ be a bag of $M_i$ instances as defined in (\ref{Bag}). For simplicity, we introduce our MI-ANFIS for the case of two rules. The generalization to an arbitrary number of rules is trivial. The MI-ANFIS with two Sugeno rules can be described as:
\begin{equation}
\begin{split}
\mathcal{R}^1(\mathbf{B}_p):\bigvee_{j=1}^{M_p}(  \; \mbox{If} \; x_{(p,j,1)} \; \mbox{is} \; A_{(1,1)} \; \mbox{and} \;  x_{(p,j,2)} \; \mbox{is} \; A_{(1,2)},\ldots,\mbox{and} \;x_{(p,j,D)} \; \mbox{is} \; A_{(1,D)}), \; \\\mbox{then} \; f_1=C(\mathbf{x}_{p1}\cdot \mathbf{b}^1,\;  \mathbf{x}_{p2}\cdot \mathbf{b}^1, \ldots, \mathbf{x}_{pM_p}\cdot \mathbf{b}^1)\\
\mathcal{R}^2(\mathbf{B}_p):\bigvee_{j=1}^{M_p}(  \; \mbox{If} \; x_{(p,j,1)} \; \mbox{is} \; A_{(2,1)} \; \mbox{and} \;   x_{(p,j,2)}\; \mbox{is} \; A_{(2,2)},\ldots,\mbox{and} \;x_{(p,j,D)} \; \mbox{is} \; A_{(2,D)}), \; \\\mbox{then} \; f_2=C(\mathbf{x}_{p1}\cdot \mathbf{b}^2,\;  \mathbf{x}_{p2}\cdot \mathbf{b}^2, \ldots, \mathbf{x}_{pM_p}\cdot \mathbf{b}^2)
\label{eq:mianfisrule}
\end{split}
\end{equation}

Figure \ref{fig:MI-Anfis-arch} illustrates the proposed MI-ANFIS architecture. As in the traditional ANFIS, nodes at the same layer have similar functions.
We denote the output of the $i$th node in layer $l$ as $O_{(l,i)}$
\begin{description}
      \item[\textbf{Layer 1}]\quad is an adaptive layer, it calculates the degree to which an input instance satisfies a quantifier $A$. Every node evaluates the membership degree $\mu_{A_{(k,j)}}$ of an input instance, $x$, in the fuzzy set $A_{(k,j)}$. Generally, $\mu_{A_{(k,j)}}$ is a parameterized membership function (MF), for example a Gaussian MF with
          \begin{equation}
            \mu_{A_{(k,j)}}(x)=exp({\frac{-(x-c_{kj})^2}{2{\sigma_{kj}}^2}}).
            \label{eq:gmf}
          \end{equation}

In (\ref{eq:gmf}), $c_{kj}$ and $\sigma_{kj}$ are the mean and variance of the gaussian function, and are referred to as the \textbf{premise parameters}.
      \item[\textbf{Layer 2}]\quad is a fixed layer where every node computes the product of all incoming inputs. It evaluates the degree of truth of proposition instances, or simply, \quotes{truth instances}. The output of this layer is computed using:
          \begin{equation}
            O_{(2,i)}=r_{\big(\ceil[\big]{i/ M_p},i[M_p]\big)}=\prod_{j=1}^D\mu_{A_{\big(\ceil[\big]{i/ M_p},j\big)}}(x_{(p,i[M_p],j)}).
            \label{eq:layer2MI-ANFIS}
          \end{equation}
 where $\ceil[\big]{}$ is a ceiling operator, and $i[M_p]$ is $1+ ((i - 1)\;mod\;M_p )$. As in the traditional ANFIS, any T-norm can replace the product as the node function in this layer.

      \item[\textbf{Layer 3}]\quad is a new addition when compared to the traditional ANFIS. Every node in this layer aggregates the truth instances (within each bag) of the previous layer by means of a smooth T-conorm. In this paper, we use a \quotes{softmax} function ($\mathcal{S}_{\alpha}$):
          \begin{equation}
            softmax_{\alpha}(x_{1}, x_{2}, \ldots, x_{n}) = \mathcal{S}_{\alpha}(x_{1}, x_{2}, \ldots, x_{n}) = \sum_{i=1}^n \frac{x_i\cdot e^{\alpha x_i}}{\sum_{j=1}^n e^{\alpha x_j}}.
            \label{eq:smax}
          \end{equation}

          In (\ref{eq:smax}), $\alpha$ determines the behavior of softmax. As $\alpha$ approaches $\infty$, softmax approaches the max operator. When $\alpha =0$, it calculates the mean. As $\alpha$ approaches $-\infty$, softmax approaches the min operator. The outputs of this layer are the firing strength of each input bag in each multiple instance fuzzy rule. i.e.,
          \begin{equation}
            O_{(3,i)}=w_i=\mathcal{S}_{\alpha}(\{r_{(i,j)}\}_{j=1}^{M_p}).
            \label{eq:layer3MI-ANFIS}
          \end{equation}
          Layer $3$ is also a fixed layer.

      \item[\textbf{Layer 4}]\quad is a fixed layer. Every node in this layer calculates the \textbf{normalized firing strength} of each rule, i.e.,
          \begin{equation}
            O_{(4,i)}=\overline{w}_i = \frac{w_i}{\sum_{j=1}^{|O_3|}w_j}.
            \label{eq:layer4MI-ANFIS}
          \end{equation}
          where $|O_3|$ is the number of rules.
      \item[\textbf{Layer 5}]\quad is an adaptive layer. Every node $i$ in this layer computes the output of the $i^{th}$ multiple instance rule using
          \begin{equation}
            O_{(5,i)}=\overline{w}_iC(\mathbf{x}_{p1}\cdot \mathbf{b}^i,\;  \mathbf{x}_{p2}\cdot \mathbf{b}^i, \ldots, \mathbf{x}_{pM_p}\cdot \mathbf{b}^i).
            \label{eq:layer5MI-ANFIS}
          \end{equation}
         The parameters $\{\mathbf{b}^i\}_{i=1}^{|O_3|}$ are referred to as the \textbf{consequent parameters}.
          The only constraint on $C$ is that it has to be a smooth function to allow for optimization techniques to be applied. We use the \quotes{softmax} as the combining function for this layer. In this case, (\ref{eq:layer5MI-ANFIS}) is equivalent to:
          \begin{equation}
            O_{(5,i)}=\overline{w}_i\mathcal{S}_{\alpha}(\mathbf{x}_{p1}\cdot \mathbf{b}^i,\;  \mathbf{x}_{p2}\cdot \mathbf{b}^i, \ldots, \mathbf{x}_{pM_p}\cdot \mathbf{b}^i).
            \label{eq:layer5MI-ANFIS1}
          \end{equation}
          note that the constant $\alpha$ here is not necessary the same as in Layer $3$.
      \item[\textbf{Layer 6}]\quad is a fixed layer with a single node labeled $\Sigma$. It computes the overall output of the system using
          \begin{equation}
            O_{(6,1)}=\sum_{i=1}^{|O_3|}O_{5,i}=\sum_{i=1}^{|O_3|}\overline{w}_i\mathcal{S}_{\alpha}(\mathbf{x}_{p1}\cdot \mathbf{b}^i,\;  \mathbf{x}_{p2}\cdot \mathbf{b}^i, \ldots, \mathbf{x}_{pM_p}\cdot \mathbf{b}^i).
            \label{eq:layer6MI-ANFIS}
          \end{equation}
    \end{description}

To learn the parameters of the proposed MI-ANFIS network, we propose a generalization to the basic learning algorithm presented by Jang \cite{jang1993anfis}. Our variation is different from the ANFIS standard backpropagation learning rule due to the additional layers (Layers 3 and 5) and the use of \quotes{softmax} function (in (\ref{eq:layer3MI-ANFIS}) and (\ref{eq:layer5MI-ANFIS})). Thus, all update equations need to be rederived.
\\\textbf{BackPropagation Learning Rule:}\label{MI-ANFIS-Backward}
we assume that we have $N$ training bags, $\mathcal{B}=\{\mathbf{B}_p \; | \; p=1,\ldots,N\}$, and their corresponding labels $\mathcal{T}=\{t_p \;|\; p=1,\ldots,N\}$. After presenting the $p$th training bag, we compute its squared error measure: 
\begin{equation}
E_p=(t_p-O_p)^2.
\label{eq:bpEp}
\end{equation}
In (\ref{eq:bpEp}), $t_p$ is the desired bag output, and $O_p$ is the computed output of the network when presented with training bag $B_p$. Recall that labels at the instances level are not available and errors can be computed only at the bag level.\\
The overall error measure of the network after presenting all $N$ bags is
\begin{equation}
E=\sum_{p=1}^N E_p.
\label{eq:bpE}
\end{equation}

To develop the gradient descent optimization on E, we compute the error rate for the $p$th training bag at each output node $O_{(l,i)}$. This error rate $\varepsilon_{(l,i)}$ (where $ 1\leq l\leq6$ indicates the MI-ANFIS layer) is defined as
\begin{equation}
\varepsilon_{l,i}=\frac{\partial E_p}{\partial O_{(l,i)}}.
\label{eq:bpepsilon}
\end{equation}
At the output node, we have
\begin{equation}
\varepsilon_{(6,1)}=\frac{\partial E_p}{\partial O_{(6,1)}}=\frac{\partial E_p}{\partial O_p}=-2(t_p-O_p).
\label{eq:bpepsilon6}
\end{equation}
For non-output nodes (i.e. internal nodes, $l<6$), we derive the error rate using the chain rule
\begin{equation}
\varepsilon_{(l,i)}=\frac{\partial E_p}{\partial O_{(l,i)}}=\sum_{h=1}^{Card(l+1)}\frac{\partial E_p}{\partial O_{(l+1,h)}}\frac{\partial O_{(l+1,h})}{\partial O_{(l,i)}}.
\label{eq:bpepsilonothernodes}
\end{equation}
where $Card(l+1)$ refers the number of nodes at layer $l+1$.
\\Next, we seek to minimize the network error with respect to the premise parameters $\{c_{kj}, \sigma_{kj} \; | \; 1\leq k \leq |O_3|, 1\leq j \leq D\}$, and with respect to the consequent parameters $\{\mathbf{b}^i\}_{i=1}^{|O_3|}$.
\\The error rate with respect to a generic parameter $\theta$ can be computed using
\begin{equation}
\frac{\partial E_p}{\partial \theta}=\sum_{O^* \in G}\frac{\partial E_p}{\partial O^*}\frac{\partial O^*}{\partial \theta}.
\label{eq:bpepsilontheta}
\end{equation}
where $G$ is the set of nodes whose outputs depend on $\theta$.
\\Using(\ref{eq:bpE}), the total error rate is given by
\begin{equation}
\frac{\partial E}{\partial \theta}=\sum_{p=1}^N\frac{\partial E_p}{\partial \theta}.
\label{eq:bpepsilonthetaall}
\end{equation}
\\\textbf{Update Rule For Premise Parameters:}\label{MI-ANFIS-BPPrem}
First we compute the error rate for the premise parameters $c_{kj}$ and $\sigma_{kj}$ using
\begin{equation}
\frac{\partial E_p}{\partial c_{kj}}=\sum_{i=1}^{M_p}\frac{\partial E_p}{\partial O_{(1,i+[(k-1)D+(j-1)]M_p)}}\frac{\partial O_{(1,i+[(k-1)D+(j-1)]M_p)}}{\partial c_{kj}}.
\label{eq:bpepsilonthetac}
\end{equation}
and,
\begin{equation}
\frac{\partial E_p}{\partial \sigma_{kj}}=\sum_{i=1}^{M_p}\frac{\partial E_p}{\partial O_{(1,i+[(k-1)D+(j-1)]M_p)}}\frac{\partial O_{(1,i+[(k-1)D+(j-1)]M_p)}}{\partial \sigma_{kj}}.
\label{eq:bpepsilonthetasigma}
\end{equation}
Using the chain rule defined in (\ref{eq:bpepsilonothernodes}), it can be shown that (see derivation in Appendix \ref{updateeq1})
\begin{equation}
\begin{split}
\frac{\partial E_p}{\partial c_{kj}}& = -2(t_p-O_p)\times \mathcal{S}_{\alpha}(\mathbf{x}_{p1}\cdot \mathbf{b}^k,\;  \mathbf{x}_{p2}\cdot \mathbf{b}^k, \ldots, \mathbf{x}_{pM_p}\cdot \mathbf{b}^k) \times \frac{ \sum_{l=1}^{|O_3|}w_l - w_k}{\Big(\sum_{l=1}^{|O_3|}w_l\Big)^2}\\
& \times \sum_{i=1}^{M_p} \Bigg(\frac{e^{\alpha r_{(k,(i+(k-1)M_p))}}}{\sum_{m=1}^{M_p} e^{\alpha r_{(k,m)}}}\Big[1+\alpha\Big(r_{(k,(i+(k-1)M_p))}-\mathcal{S}_{\alpha}(\{r_{(k,m)}\}_{m=1}^{M_p})\Big)\Big] \\ & \times \prod_{d=1, d\neq j}^D\mu_{A_{\Big(\ceil[\big]{(i+(k-1)M_p)/ M_p},d\Big)}}\Big(x_{(p,(i+(k-1)M_p)[M_p],d)}\Big)
\\ & \times \frac{(x_{(p,(i+(k-1)M_p)[M_p],j)}-c_{kj})}{\sigma_{kj}^2}\times exp({-\frac{(x_{(p,(i+(k-1)M_p)[M_p],j)}-c_{kj})^2}{2\sigma_{kj}^2}}) \Bigg).
\label{eq:bpepsilonthetacval}
\end{split}
\end{equation}
\\The center parameters $c_{kj}$ are then updated using
\begin{equation}
\triangle c_{kj}=-\eta \frac{\partial E}{\partial c_{kj}}.
\label{eq:updateformulackj}
\end{equation}
where $\eta$ is the learning rate.

The update formula for $\sigma_{kj}$ can be derived in a similar manner. It can be shown that
\begin{equation}
\begin{split}
\frac{\partial E_p}{\partial \sigma_{kj}}& = -2(t_p-O_p)\times \mathcal{S}_{\alpha}(\mathbf{x}_{p1}\cdot \mathbf{b}^k,\;  \mathbf{x}_{p2}\cdot \mathbf{b}^k, \ldots, \mathbf{x}_{pM_p}\cdot \mathbf{b}^k) \times \frac{ \sum_{l=1}^{|O_3|}w_l - w_k}{\Big(\sum_{l=1}^{|O_3|}w_l\Big)^2}\\
& \times \sum_{i=1}^{M_p} \Bigg(\frac{e^{\alpha r_{(k,(i+(k-1)M_p))}}}{\sum_{m=1}^{M_p} e^{\alpha r_{(k,m)}}}\Big[1+\alpha\Big(r_{(k,(i+(k-1)M_p))}-\mathcal{S}_{\alpha}(\{r_{(k,m)}\}_{m=1}^{M_p})\Big)\Big] \\ & \times \prod_{d=1, d\neq j}^D\mu_{A_{\Big(\ceil[\big]{(i+(k-1)M_p)/ M_p},d\Big)}}\Big(x_{(p,(i+(k-1)M_p)[M_p],d)}\Big)
\\ & \times \frac{(x_{(p,(i+(k-1)M_p)[M_p],j)}-c_{kj})^2}{\sigma_{kj}^3}\times exp({-\frac{(x_{(p,(i+(k-1)M_p)[M_p],j)}-c_{kj})^2}{2\sigma_{kj}^2}}) \Bigg).
\label{eq:bpepsilonthetacval}
\end{split}
\end{equation}
The MF's width, $\sigma_{kj}$, are then updated using
\begin{equation}
\triangle \sigma_{kj}=-\eta \frac{\partial E}{\partial \sigma_{kj}}.
\label{eq:updateformulasigmakj}
\end{equation}
\\\textbf{Update Rule For Consequent Parameters:}\label{MI-ANFIS-BPPrem}
The error rate for the consequent parameters $\{\mathbf{b}^i=\{b_0^i,...,b_D^i\}, i=1\ldots |O_3|\}$ is defined as
\begin{equation}
\frac{\partial E_p}{\partial \mathbf{b}^i}=\big(\frac{\partial E_p}{\partial b_0^i},\frac{\partial E_p}{\partial  b_1^i},\ldots,\frac{\partial E_p}{\partial  b_D^i}\big).
\label{eq:bpepsilonthetconsBall}
\end{equation}
where,
\begin{equation}
\frac{\partial E_p}{\partial b_j^i}=\frac{\partial E_p}{\partial O_{(5,i)}}\frac{\partial O_{(5,i)}}{\partial b_j^i}, \;for\;\; j=1,\ldots,D.
\label{eq:bpepsilonthetconsB}
\end{equation}
Using (\ref{eq:bpepsilonothernodes}), it can be shown that (see Appendix \ref{updateeq2})
\begin{multline}
\frac{\partial E}{\partial b_{j}^i} =\sum_{p=1}^N\frac{\partial E_p}{\partial b_{j}^i}
 =\sum_{p=1}^N\overline{w}_i\sum_{m=1}^{M_p}\Bigg(\frac{1}{\Big(\sum_{h=1}^{M_p} exp(\alpha(\mathbf{x}_{ph}\cdot \mathbf{b}^i-\mathbf{x}_{pm}\cdot \mathbf{b}^i))\Big)^2}\\\times\Big[\Big(x_{(p,m,j)}\sum_{h=1}^{M_p} exp(\alpha(\mathbf{x}_{ph}\cdot \mathbf{b}^i-\mathbf{x}_{pm}\cdot \mathbf{b}^i)\Big)
 -\Big(\mathbf{x}_{pm}\cdot \mathbf{b}^i\sum_{h=1}^{M_p} exp(\alpha(\mathbf{x}_{ph}\cdot \mathbf{b}^i-\mathbf{x}_{pm}\cdot \mathbf{b}^i)\alpha(x_{(p,h,j)}-x_{(p,m,j)})\Big)\Big]\Bigg).
\label{eq:bpepsilonthetaallbij}
\end{multline}
The consequent parameters are then updated using
\begin{equation}
\triangle b_{j}^i=-\eta \frac{\partial E}{\partial b_{j}^i}.
\label{eq:updateformulabij}
\end{equation}
Equations (\ref{eq:updateformulackj}), (\ref{eq:updateformulasigmakj}) and (\ref{eq:updateformulabij}) can be used to update $c_{kj}$, $\sigma_{kj}$ and $b_{j}^i$ parameters either on-line, bag by bag ( we want to emphasis here that the on-line learning is not achieved instance by instance, but rather bag by bag), or off-line in batch mode after presentation of the entire data.

The proposed MI-ANFIS learning algorithm is summarized in Algorithm \ref{alg1}.
\begin{algorithm}
  \caption{MI-ANFIS Basic Learning Algorithm}
  \label{alg1}
  {\begin{tabbing}
  \textbf{Inputs}: \=$\mathcal{B}$: the set of training bags.\\
   \>$\mathcal{T}$: labels of the training bags.\\
   \>$M$: the number of instances in each bag.\\
   \>$\alpha$: the constant used in the \quotes{softmax} function.\\
   \>$\eta$: the learning rate.\\
   \>$E_{max}$: number of epochs.\\
   \>$\epsilon$: minimum parameters change value.\\
   \textbf{Outputs}: \=$\mathbf{b}^i$: the sets of consequent parameters.\\
     \>$\mathbf{c}^i$: the set of membership functions' centers.\\
   \>$\mathbf{\sigma}^i$: the set of membership functions' widths.
   \end{tabbing}}
  \begin{algorithmic}
    \STATE Initialize $\mathbf{b}^i$, $\mathbf{c}^i$, and $\mathbf{\sigma}^i$.
    \REPEAT
       \STATE Update $\mathbf{b}^i$ using (\ref{eq:updateformulabij}) and $b^{i(new)}=b^{i(old)}+\triangle b^i$.
       \STATE Update $\mathbf{c}^i$ using (\ref{eq:updateformulackj}) and  $c^{i(new)}=c^{i(old)}+\triangle c^i$.
       \STATE Update $\mathbf{\sigma}^i$ using (\ref{eq:updateformulasigmakj}) and $\sigma^{i(new)}=\sigma^{i(old)}+\triangle \sigma^i$.

    \UNTIL{\small{$max(\|b^{i(new)}-b^{i(old)}\|,\|c^{i(new)}-c^{i(old)}\|,\|\sigma^{i(new)}-\sigma^{i(old)}\|)< \epsilon \; \; or\;\; Number \;of\; epochs\; > E_{max}$}}
    \RETURN $\mathbf{b}^i$, $\mathbf{c}^i$, $\mathbf{\sigma}^i$
  \end{algorithmic}
\end{algorithm}
\section{Preventing Overfitting: Rule Dropout}\label{MI-ANFIS-Overfit}
\begin{figure}[htb]
  \centering
  \includegraphics[width=0.8\linewidth]{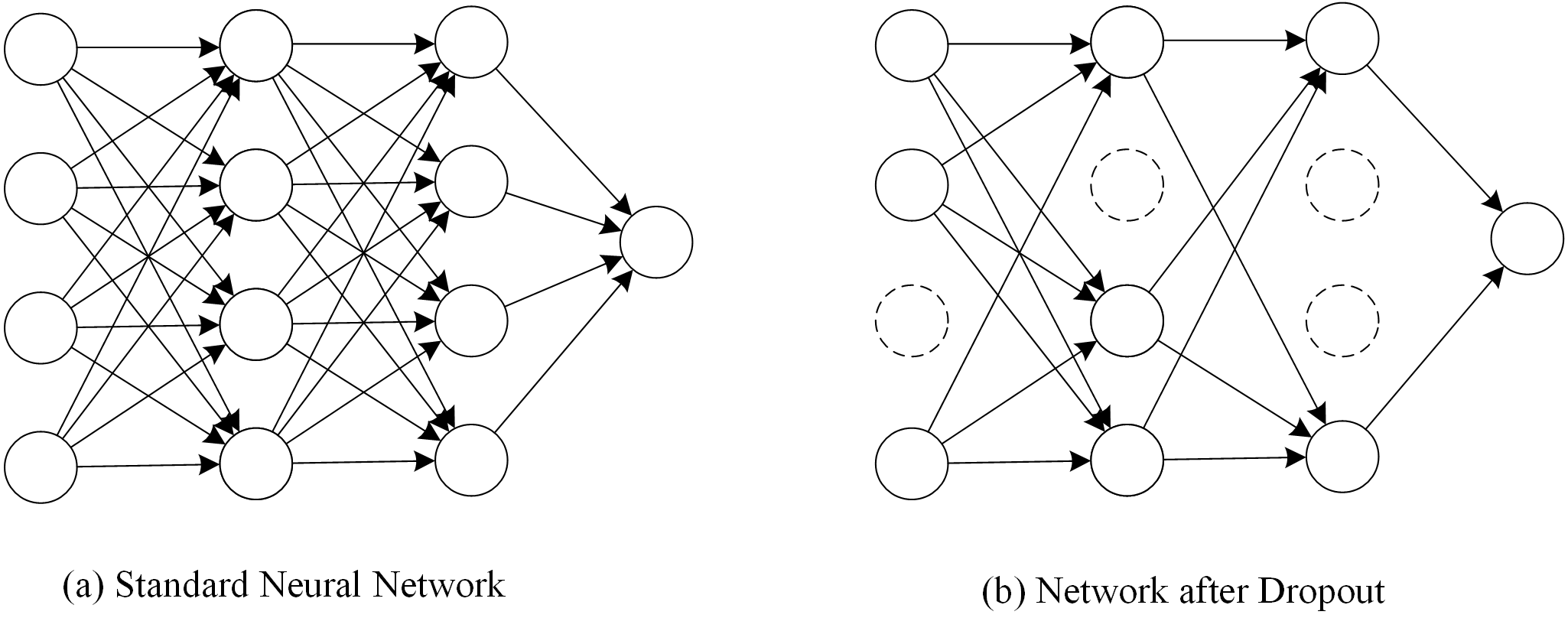}
   \caption[Dropout neural network model]{Dropout neural network model. (a) is a standard neural network. (b) is the same network after applying dropout. Doted lines indicate a node that has been dropped. (source \cite{srivastava2014dropout})}
     \label{fig:Dropout2}
\end{figure}

\begin{figure}[htb]
  \centering
  \includegraphics[width=0.8\linewidth]{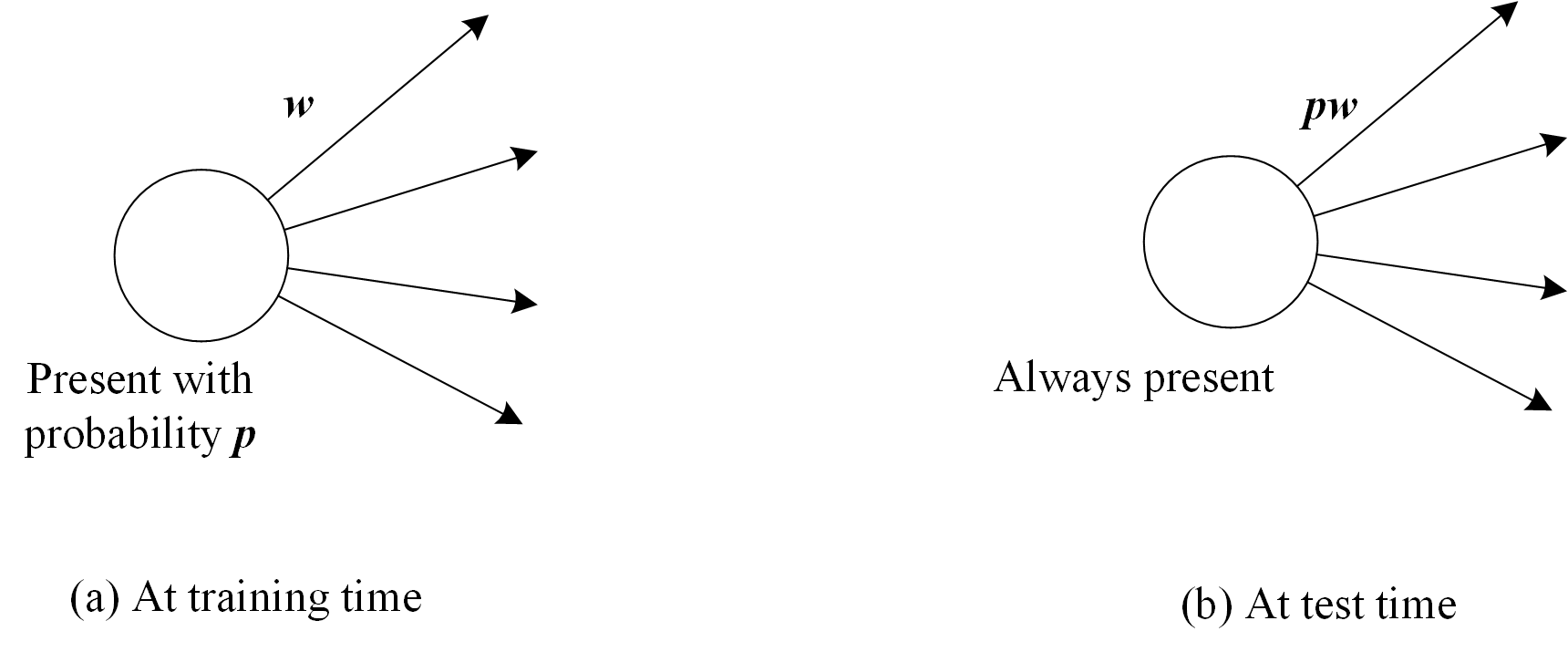}
   \caption[Illustration of Dropout application.]{Illustration of Dropout application. (a) a node is dropped with probability $1-p$ at training time. (b) at test time the node is always present and its outputs are weighted by $p$. (source \cite{srivastava2014dropout})}
     \label{fig:Dropout1}
\end{figure}

Neural networks with large number of parameters are susceptible to overfitting. MI-ANFIS is no exception, particularly when using large number of multiple instance fuzzy rules and relatively small training datasets. In such scenario, some rules could co-adapt to the training data and degrade the network ability to generalize to unseen examples. In this section, we present a technique, known as Dropout, used to prevent overfitting and rules' co-adaptation.
\\Dropout is a regularization method that was introduced by Hinton et al. \cite{srivastava2014dropout} to alleviate the serious problem of overfitting in deep neural networks. Over the years, many methods have been developed to reduce overfitting, including using a validation dataset to stop the training as soon as the performance gets worse, adding weight penalties using L1 and L2 regularization, or artificially augmenting the training dataset using label-preserving transformations. However, as noted by Hinton \cite{srivastava2014dropout}, the best way to regularize a fixed-size model is to average the predictions of all possible settings of the parameters weighted by its posterior probability given the training data. This can be achieved by combining the predictions of an exponential number of models. Combining several models with different architectures have the advantage of better generalization and per consequence better testing performance. While generating an ensemble of models is trivial, training them all is prohibitively expensive.
\\Generally, Dropout works by setting to $0$ the output of each node in a given layer with probability $1-\mathbf{p}$ ($p$ typically equals $0.5$), during training. Nodes that are dropped out do not contribute to the parameters updates. During testing, all nodes are used but the outputs are weighted by the probability $\mathbf{p}$. Following this strategy, every time a new training example is presented, the network samples and trains a different architecture. In other words, Dropout trains an ensemble of networks ($2^N$ networks, $N$ being the number of nodes) simultaneously leading to an important speedup in training time as compared to traditional ensemble methods. Figure \ref{fig:Dropout2} and Figure \ref{fig:Dropout1} illustrate the Dropout model.

\begin{figure}[htb]
  \centering
  \includegraphics[width=0.9\linewidth]{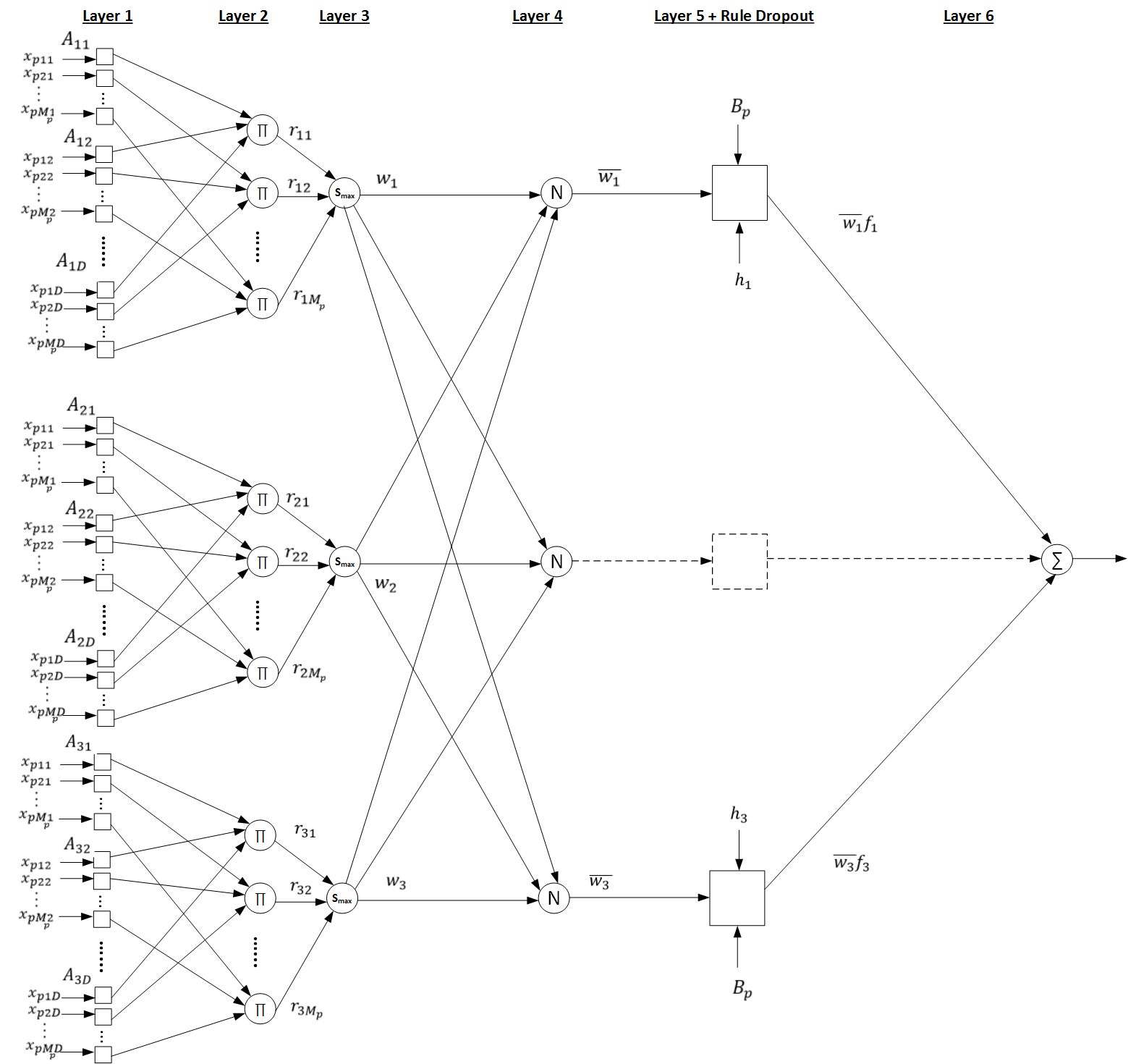}
   \caption{Illustration of Rule Dropout application. Doted lines indicate a rule that has been dropped.}
     \label{fig:MIANFISDropout2}
\end{figure}

In this paper, we propose to adopt the Dropout strategy to regularize MI-ANFIS networks. Typically, overfitting occurs in MI-ANFIS networks when a set of multiple instance rules co-adapt to the provided data early during the training process and prevent the remaining rules from learning. Thus, degrading the network's generalization capability. While the Dropout technique could be applied to MI-ANFIS as is (given the inherited neural network nature of the architecture), care should be exercised when selecting nodes to include in the list of the randomly dropped out nodes. MI-ANFIS nodes are different from that of standard neural networks as they are grouped into rules to model and express linguistic terms. Simply dropping few nodes from a given rule can change its role and could severely handicap the fuzzy inference process. Hence, Dropout should be executed differently. In deep neural nets, Dropout is applied to selected layers (vertically), for MI-ANFIS, we propose to apply Dropout on a rule by rule basis (i.e., horizontally). Either the whole rule is included, or the whole rule is dropped. This can be achieved by applying Dropout to Layer $5$ (see Figure \ref{fig:MIANFISDropout2}), i.e., setting to zero the output of the \quotes{to be dropped out} rules. We will refer to this derived technique as \quotes{\textbf{Rule Dropout}}. Using a Rule Dropout strategy to train MI-ANFIS networks is approximatively equivalent to sampling and training $2^R$ ($R$ is the number of rules) ensemble of networks.

Let $p$ be the probability with which a rule is present, formally, Rule Dropout is applied to Layer $5$ during training as follows

\begin{equation}
O_{5,i}=h_i\overline{w}_i\mathcal{S}_{\alpha}(\mathbf{x}_{p1}\cdot \mathbf{b}^i,\;  \mathbf{x}_{p2}\cdot \mathbf{b}^i, \ldots, \mathbf{x}_{pM_p}\cdot \mathbf{b}^i),
\label{eq:layer5MI-ANFIS1Drop2}
\end{equation}
where
\begin{equation}
h_i \thicksim Bernoulli(p)
\label{eq:layer5MI-ANFISDrop1}
\end{equation}
is a Bernoulli random variable with probability $p$ of being $1$. During testing, Layer $5$ output is scaled by $p$, i.e., $O_{3,i}=p\overline{w}_i\mathcal{S}_{\alpha}(\mathbf{x}_{p1}\cdot \mathbf{b}^i,\;  \mathbf{x}_{p2}\cdot \mathbf{b}^i, \ldots, \mathbf{x}_{pM_p}\cdot \mathbf{b}^i)$. Figure \ref{fig:MIANFISDropout2} illustrates our MI-ANFIS network with 3 multiple instance fuzzy rules where, at a given iteration, rule 2 has been dropped out..

Deriving the new update equations for MI-ANFIS parameters requires taking into consideration the added Bernoulli random variable, $h_i$. It is straightforward to show that the new gradients with respect to premise and consequent parameters are given by
{\small
\begin{multline}
\frac{\partial E_p}{\partial c_{kj}} = -2(t_p-O_p)\\\times h_k\times\mathcal{S}_{\alpha}(\mathbf{x}_{p1}\cdot \mathbf{b}^k,\;  \mathbf{x}_{p2}\cdot \mathbf{b}^k, \ldots, \mathbf{x}_{pM_p}\cdot \mathbf{b}^k) \times \frac{ \sum_{l=1}^{|O_3|}w_l - w_k}{\Big(\sum_{l=1}^{|O_3|}w_l\Big)^2}\\
 \times \sum_{i=1}^{M_p} \Bigg(\frac{e^{\alpha r_{k,(i+(k-1)M_p)}}}{\sum_{m=1}^{M_p} e^{\alpha r_{k,m}}}\Big[1+\alpha\Big(r_{k,(i+(k-1)M_p)}-\mathcal{S}_{\alpha}(\{r_{k,m}\}_{m=1}^{M_p})\Big)\Big] \\
  \times \prod_{d=1, d\neq j}^D\mu_{A_{\Big(\ceil[\big]{(i+(k-1)M_p)/ M_p},d\Big)}}\Big(x_{p,(i+(k-1)M_p)[M_p],d}\Big)
\\  \times \frac{(x_{(p,(i+(k-1)M_p)[M_p],j)}-c_{kj})}{\sigma_{kj}^2}\\\times exp({-\frac{(x_{(p,(i+(k-1)M_p)[M_p],j)}-c_{kj})^2}{2\sigma_{kj}^2}}) \Bigg).
\label{eq:bpepsilonthetacval1RD}
\end{multline}
}
and,
{\small
\begin{multline}
\frac{\partial E_p}{\partial \sigma_{kj}} = -2(t_p-O_p)\\\times h_k\times\mathcal{S}_{\alpha}(\mathbf{x}_{p1}\cdot \mathbf{b}^k,\;  \mathbf{x}_{p2}\cdot \mathbf{b}^k, \ldots, \mathbf{x}_{pM_p}\cdot \mathbf{b}^k) \times \frac{ \sum_{l=1}^{|O_3|}w_l - w_k}{\Big(\sum_{l=1}^{|O_3|}w_l\Big)^2}\\
 \times \sum_{i=1}^{M_p} \Bigg(\frac{e^{\alpha r_{k,(i+(k-1)M_p)}}}{\sum_{m=1}^{M_p} e^{\alpha r_{k,m}}}\Big[1+\alpha\Big(r_{k,(i+(k-1)M_p)}-\mathcal{S}_{\alpha}(\{r_{k,m}\}_{m=1}^{M_p})\Big)\Big] \\
  \times \prod_{d=1, d\neq j}^D\mu_{A_{\Big(\ceil[\big]{(i+(k-1)M_p)/ M_p},d\Big)}}\Big(x_{p,(i+(k-1)M_p)[M_p],d}\Big)
\\ \times \frac{(x_{(p,(i+(k-1)M_p)[M_p],j)}-c_{kj})^2}{\sigma_{kj}^3}\\\times exp({-\frac{(x_{(p,(i+(k-1)M_p)[M_p],j)}-c_{kj})^2}{2\sigma_{kj}^2}}) \Bigg).
\label{eq:bpepsilonthetacvalRD}
\end{multline}
}
In a similar manner,
{\small
\begin{equation}
\begin{split}
\frac{\partial E}{\partial b_{j}^i} &=\sum_{p=1}^N -2(t_p-O_p)\\&\times h_i\overline{w}_i\sum_{m=1}^{M_p}\frac{1}{\Big(\sum_{h=1}^{M_p} exp(\alpha(\mathbf{x}_{ph}\cdot \mathbf{b}^i-\mathbf{x}_{pm}\cdot \mathbf{b}^i))\Big)^2}\\&\times\Big[\Big(x_{pmj}\sum_{h=1}^{M_p} exp(\alpha(\mathbf{x}_{ph}\cdot \mathbf{b}^i-\mathbf{x}_{pm}\cdot \mathbf{b}^i)\Big)
\\ & -\Big(\mathbf{x}_{pm}\cdot \mathbf{b}^i\sum_{h=1}^{M_p} exp(\alpha(\mathbf{x}_{ph}\cdot \mathbf{b}^i-\mathbf{x}_{pm}\cdot \mathbf{b}^i)\alpha(x_{phj}-x_{pmj})\Big)\Big].
\end{split}
\label{eq:bpepsilonthetaallbijRD}
\end{equation}
}
As it can be seen, equations (\ref{eq:bpepsilonthetacval1RD}), (\ref{eq:bpepsilonthetacvalRD}), and (\ref{eq:bpepsilonthetaallbijRD}) will get zeroed when the rule is dropped out (i.e., $h_k=0$ and $h_i=0$). Thus, its premise and consequent parameters are not updated.

\section{EXPERIMENTAL RESULTS}\label{SyntheticData}
\subsection{Synthetic Data}
\begin{figure}[htb]
  \centering
  \includegraphics[width=0.9\linewidth]{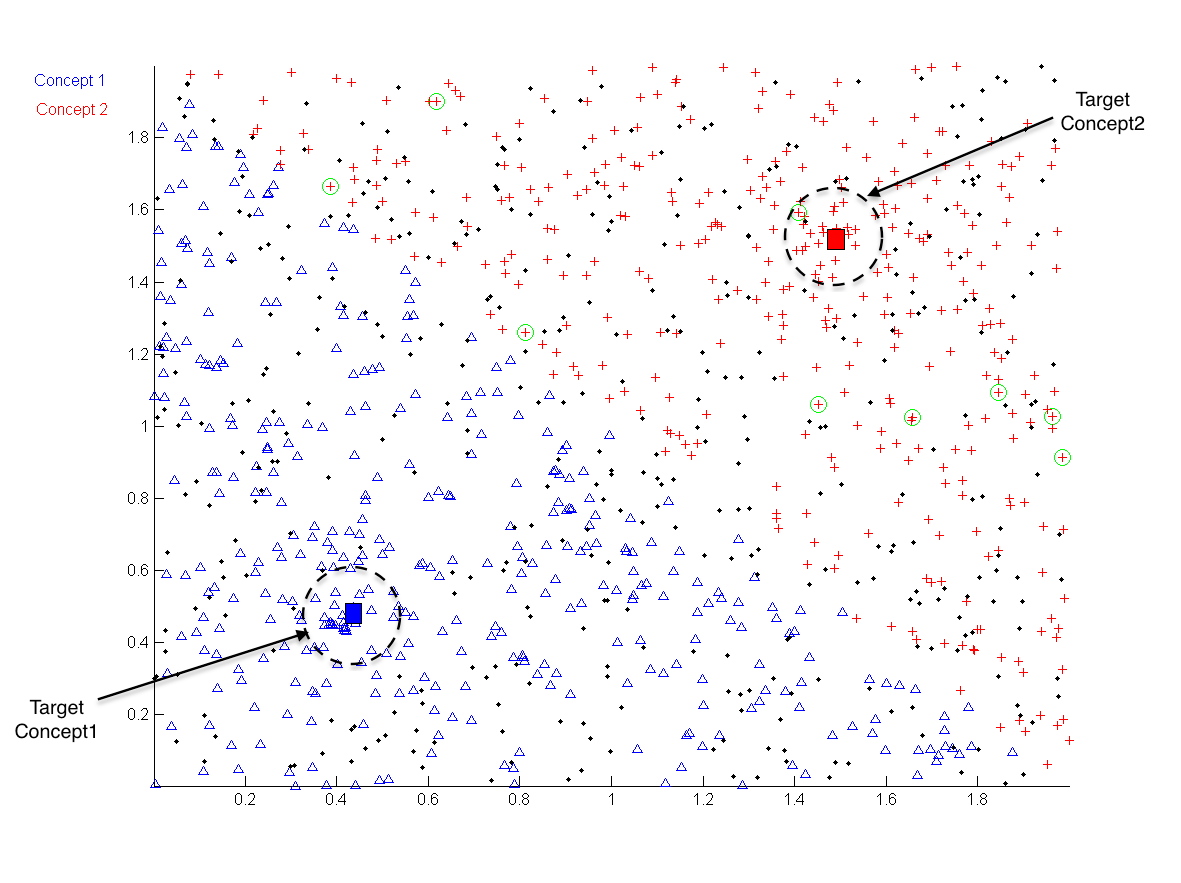}
   \caption{Instances from positive and negative bags drawn from data that have 2 concepts. The center of each target concept is indicated by a square and the circles indicate the region within 1 - standard deviation from the mean. Instances from negative bags are shown as ``.'', and instances from positive bags are shown as ``+'' or ``$\vartriangle$''. Instances from one sample positive bag are circled.}
  \label{fig:Datasimple2bags}
\end{figure}

To illustrate the proposed multiple instance fuzzy inference and its ability to learn from data without instance-level labels, first, we use a simple 2-Dim synthetic dataset. This data were generated from a distribution of two positive contexts with centers at (0.5,0.5) and (1.5,1.5), and with a fixed standard deviation. These centers are marked with squares in Figure \ref{fig:Datasimple2bags}, and the circles around the centers indicates regions within 1 standard deviation. These regions are considered the two target concepts. From each positive concept we generated 50 bags. Each bag has a random number, between 2 and 10, of instances. Each bag from concept 1 (or 2) will have at least one instance close to target concept 1 (or 2). We also generated 50 negative bags randomly from non concept regions. Negative bags will have all of their instances outside both target concepts. In Figure \ref{fig:Datasimple2bags}, instances from negative bags are shown as ``.'', and instances from positive bags are shown as ``+'' or ``$\vartriangle$'' depending on the underlying concept. In Figure \ref{fig:Datasimple2bags}, we highlight one bag from Concept 1 by circling all of its instances. As it can be seen, one of its instances is within one standard deviation region of target concept 1 while the other instances are scattered around. We should emphasize here that the centers of the target concepts in Figure \ref{fig:Datasimple2bags} are unknown and not used by the learning algorithm. They are shown here for illustration and validation purposes only.
\subsubsection{MI-ANFIS Rules Learning}
 In the following, we show that the MI-ANFIS Learning Algorithm (Algorithm \ref{alg1}) is capable of identifying positive concepts as well as their corresponding multiple instance fuzzy rules.
To initialize the premise parameters, we partition the instances' space into 6 partition generated randomly \footnote{A grid or manual partitioning could also be used}. We use the partitions' centers as initial centers for the Gaussian MFs, and we initialize all standard deviation parameters to a default value of $0.5$.

\begin{figure}[h!]
  \centering
  \includegraphics[width=0.9\linewidth]{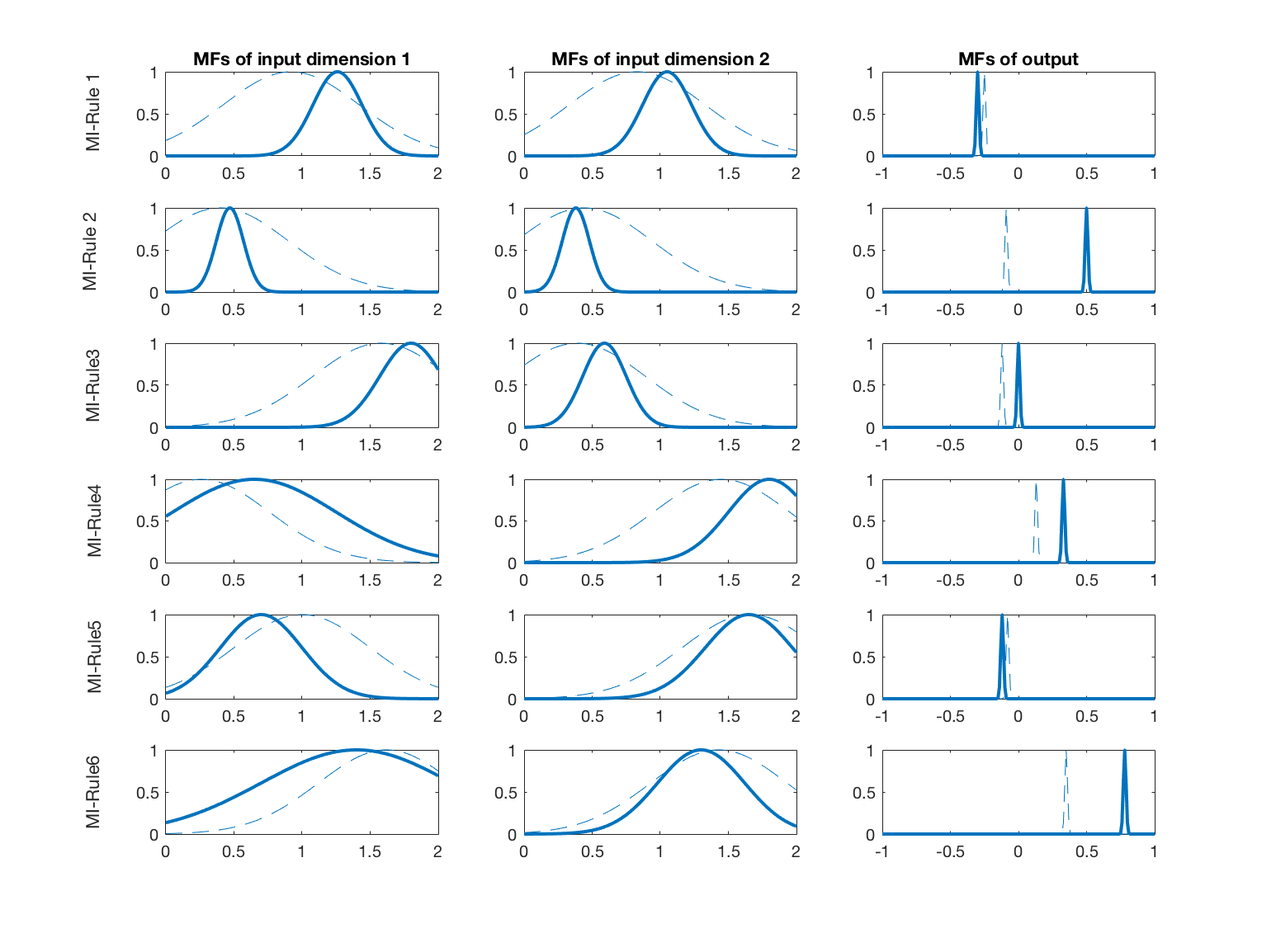}
   \caption{Learned MFs after convergence of MI-ANFIS training algorithm. Initial MFs before training are marked with dashed lines. Learned MFs are shown in sold bold lines.}
  \label{fig:synth}
\end{figure}

The initial fuzzy sets (MFs) of the rules, before training, are displayed in Figure \ref{fig:synth} in dashed lines. As it can be seen, the initial 6 partitions simply cover random quadrants of the 2D instance space (if no label information is used, as in this case, data would appear to have uniform distribution (refer to Figure \ref{fig:Datasimple2bags})). The learned fuzzy sets after convergence are shown in Figure \ref{fig:synth} in bold lines. As it can be seen, 
%
the system has correctly identified the positive concepts, and at the same time identified irrelevant rules (MI-Rule 1, MI-Rule 3 and MI-Rule 5) and assigned low output values to each, $-0.3$, $-0.06$ and $-0.12$ respectively. 

\subsection{Benchmark Datasets}
To provide a quantitative evaluation of the proposed MI-ANFIS, we apply it to five benchmark data sets commonly used to evaluate MIL methods: The MUSK1, MUSK2 \cite{dietterich1997solving}, and Fox, Tiger, and Elephant from the COREL data set \cite{andrews2002support}. MUSK1 and MUSK2 data sets consist of descriptions of molecules and the objective is to classify whether a molecule smells musky \cite{li2013multiple}. In these data sets, each bag represents a molecule and instances within each bag represent the different low-energy conformations of the molecule. Each instance is characterized by 166 features. MUSK1 has 92 bags, of which 47 are positive, and MUSK2 has 102 bags, of which 39 are positive.
The other data sets (Fox, Tiger, and Elephant), classify whether an image contains the corresponding animal. Each data set consists of 200 images (bags): 100 positive images containing the target animal and 100 negative images containing other animals. Each image is represented as a set of patches (instances) and each patch is in turn represented by a 230 dimensional feature vector describing its color, texture and shape information. We note that the three data sets are independent and used as binary classification problem (positive v.s. negative). Table \ref{data} summarizes the characteristics of the five data sets. It is to be noted that for each benchmark data set, PCA was applied to reduce the dimensionality of the features in order to speedup MI-ANFIS training and increase the interpretability of the generated multiple instance fuzzy rules.
\begin{table}[!t]
\renewcommand{\arraystretch}{1.3}
\caption{Benchmark data sets}
\label{data}
\centering
\begin{tabular}{l||c|c|c|c|c}
\hline
\bfseries Data set & \bfseries dim.(PCA) & \bfseries No. Bags & \bfseries Positive & \bfseries Negative &\bfseries No.Instances\\
\hline\hline
MUSK1 & 166(25)& 92& 47& 45&  $2 \rightarrow 40$\\
\hline
MUSK2 & 166(25)& 102& 39& 63&  $1 \rightarrow 1044$\\
\hline
Fox & 230(10)& 200& 100& 100&  $2 \rightarrow 13$\\
\hline
Tiger & 230(10)& 200& 100& 100&  $1 \rightarrow 13$\\
\hline
Elephant & 230(10)& 200& 100& 100&  $2 \rightarrow 13$\\
\hline
\end{tabular}
\end{table}
\\For each experiment, we construct a zero-order MI-ANFIS with a given number of multiple instance rules. For MI-ANFIS the number of rules is not critical. It should be large enough to cover the diverse regions of the input space and the multiple concepts. If the specified number of rules is too large, some will vanish as illustrated in Figure \ref{fig:synth} for the example with the synthetic data. Also, larger number of rules leads to slower training. We use Gaussian MFs to describe the input fuzzy sets. For initialization, we use the FCM algorithm to cluster the instances of the positive bags into a number of clusters equal to the number of fuzzy rules, and we initialize MFs' centers as the clusters centers. MI-ANFIS was trained and tested using ten fold cross validation. Table \ref{MIANFISparms} summarizes all parameters used in training the MI-ANFIS (parameters were manually selected using trial and error). We note that the reason behind using larger standard deviations for MUSK1 and MUSK2 datasets is the higher dimensionality of these data sets. We expect the sparsity to increase with the dimensions of the feature space, so we set the standard deviations to larger values to allow the initial rules to cover the entirety of the input space.\\

\begin{table}[!t]
\renewcommand{\arraystretch}{1.3}
\caption{MI-ANFIS Training Parameters}
\label{MIANFISparms}
\centering
\begin{tabular}{l||c|c|c|c|c}
\hline
\bfseries Parameter & \bfseries MUSK1 & \bfseries MUSK2& \bfseries FOX & \bfseries Tiger &\bfseries Elephant\\
\hline\hline
No. of MI Rules & 6& 3& 2& 4&  3\\
\hline
No. of Inputs & 25& 25& 10& 10&  10\\
\hline
MF's $\sigma$ & 100& 100& 10& 10&  10\\
\hline
Output parameters & 1s& 1s& 1s& 1s&  1s\\
\hline
Softmax's $\alpha$ & 1& 1& 1& 1&  1\\
\hline
Learning rate & 0.1& 0.1& 0.1& 0.1&  0.1\\
\hline
\end{tabular}
\end{table}

First, to illustrates the advantage of using MI-ANFIS over the traditional ANFIS we compare these two algorithms on the two MUSK data sets. Since ANFIS cannot learn from ambiguously labeled data,  for the sake of comparison, we consider the naive MIL assumption where all instances from positive bags are considered positive and all instances from negative bags are considered negative. We refer to this implementation as Naive-ANFIS. The results are summarized in Table \ref{resultstableANIFSes} where the performance is reported in terms of prediction accuracy averaged over all 10 cross validation sets (\% of correct $\pm$ standard deviation). As it can be seen, MI-ANFIS outperforms Naive-ANFIS significantly. This is because inaccurately labeled instances within the positive bags were used for training the Naive-ANFIS. The difference in performance between MI-ANFIS and Naive-ANFIS is greater for MUSK1 and MUSK2 because of the greater number of instances per bag (more ambiguousity). 
\begin{table}[!t]
\renewcommand{\arraystretch}{1.3}
\caption{Comparison of MI-ANFIS prediction accuracy (in percent) to Naive-ANFIS on the benchmark data sets.}
\label{resultstableANIFSes}
\centering
\begin{tabular}{l||c|c|c|c|c}
\hline
\bfseries Algorithms & \bfseries MUSK1 & \bfseries MUSK2& \bfseries Fox & \bfseries Tiger &\bfseries Elephant\\
\hline\hline
 MI-ANFIS &  93.49&  90.58& 66.4&  84.5&  86.97\\
&  $\pm0.76$& $\pm1.31$& $\pm2.77$& $\pm0.61$& $\pm1.10$\\
\hline
Naive-ANFIS & 67.82& 79.43&  58.70& 77.70&  82.2\\
& $\pm4.04$& $\pm5.04$&  $\pm1.35$&$\pm0.83$ &  $\pm0.83$\\
\hline
\end{tabular}
\end{table}

\begin{table}[!htbp]
\renewcommand{\arraystretch}{1.3}
\caption{Comparison of MI-ANFIS prediction accuracy (in percent) to other methods on the benchmark data sets. Results for 3 top performing methods are shown in bold font. We use reported results, N/A indicated that a given algorithm was not applied to that data set.}
\label{resultstable}
\centering
\begin{tabular}{l||c|c|c|c|c}
\hline
\bfseries Algorithms & \bfseries MUSK1 & \bfseries MUSK2& \bfseries Fox & \bfseries Tiger &\bfseries Elephant\\
\hline\hline
\bfseries MI-ANFIS & \bfseries 93.49& \bfseries 90.58&\bfseries 66.4& \bfseries 84.5& \bfseries 86.97\\
&  $\pm0.76$& $\pm1.31$& $\pm2.77$& $\pm0.61$& $\pm1.10$\\
\hline
MILES \cite{chen2006miles} & 86.3& 87.7& N/A& N/A&  N/A\\
\hline
APR \cite{dietterich1997solving}& 92.4& 89.2&  N/A& N/A&  N/A\\
\hline
DD \cite{Maron} & 88.9& 82.5& N/A& N/A&  N/A\\
\hline
DD-SVM \cite{chen2004image} & 85.8& \bfseries 91.3&  N/A& N/A&  N/A\\
\hline
EM-DD \cite{zhang2001dd}& 84.8& 84.9& 56.1& 72.1&  78.3\\
\hline
Citation-KNN \cite{wang2000solving} & 92.4& 86.3&  N/A& N/A&  N/A\\
\hline
MI-SVM \cite{andrews2002support}& 77.9& 84.3& 57.8& 84.0&  81.4\\
\hline
mi-SVM \cite{andrews2002support}& 87.4& 83.6& 58.2& 78.4&  82.2\\
\hline
MI-NN \cite{ramon2000multi} & 88.0& 82.0&  N/A& N/A&  N/A\\
\hline
Bagging-APR \cite{zhou2003ensembles} & \bfseries 92.8& \bfseries 93.1&  N/A& N/A&  N/A\\
\hline
RBF-MIP \cite{zhang2006adapting}&91.3 & 90.1&  N/A& N/A&  N/A\\
& $\pm1.6$ & $\pm1.7$&  & &  \\
\hline
BP-MIP \cite{zhou2002neural}& 83.7& 80.4&  N/A& N/A&  N/A\\
\hline
RBF-Bag-Unit \cite{RBF}& 90.3& 86.6& N/A& N/A&  N/A\\
\hline
MI-kernel \cite{zhou2009multi} & 88.0& 89.3& 60.3& 84.2&  84.3\\
\hline
PPPM-kernel \cite{wang2008adaptive}& \bfseries 95.6& 81.2& 60.3& 80.2&  82.4\\
\hline
MIGraph \cite{zhou2009multi}& 90.0& 90.0& 61.2& 81.9&  \bfseries 85.1\\
\hline
miGraph \cite{zhou2009multi}& 88.9& 90.3&61.6& \bfseries 86.0& \bfseries 86.8\\
\hline
ALP-SVM \cite{gehler2007deterministic} & 86.3& 86.2& \bfseries 66.0& \bfseries 86.0&  83.5\\
\hline
MIForest \cite{leistner2010miforests}& 85.0& 82.0& \bfseries 64.0& 82.0&  84.0\\
\hline
\end{tabular}
\end{table}

Table \ref{resultstable} compares the performance of the proposed algorithm to state of the art MIL algorithms on the benchmark data sets. 

 Overall, MI-ANFIS is comparable to other MIL algorithms. In fact, on all tested data sets, MI-ANFIS ranked consistently among the top three. For MUSK1, PPPM-kernel \cite{wang2008adaptive} performed the best (95.6\%), but this algorithm did not perform as well for the other sets. For MUSK2 Bagging-APR \cite{zhou2003ensembles} achieved the best accuracy, as reported by \cite{chen2006miles}. MI-ANFIS achieved the best average performance for the Fox and Elephant data sets, and second best performance after the miGraph \cite{zhou2009multi} and ALP-SVM \cite{gehler2007deterministic} methods for the Tiger data set.

\begin{figure}[!h]
  \centering
  \includegraphics[width=1\linewidth]{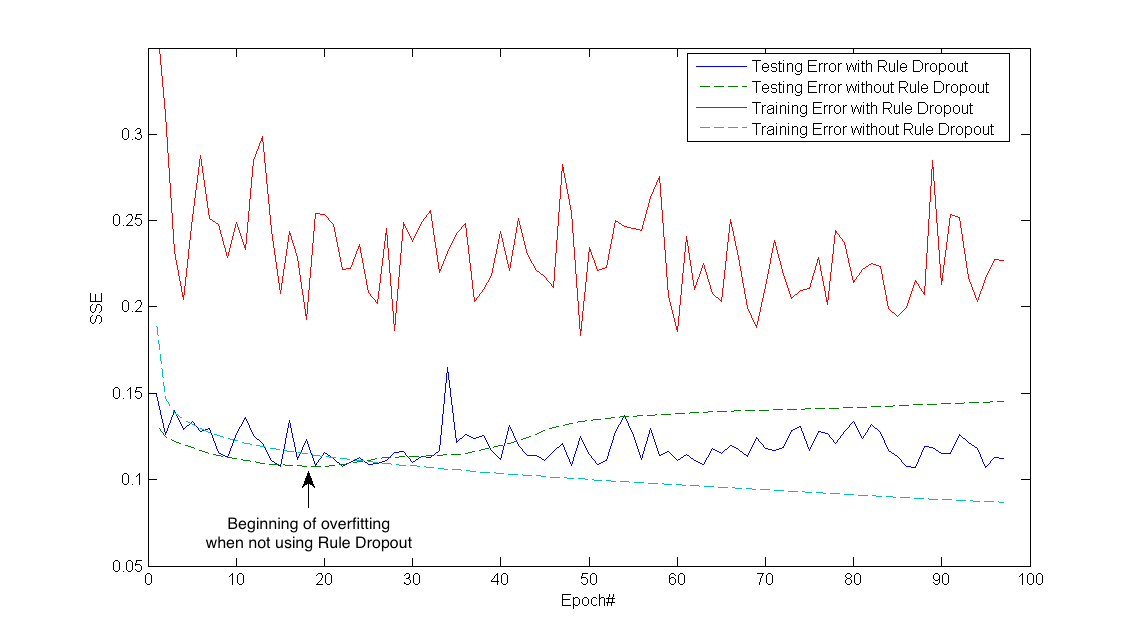}
   \caption{Training and testing errors for two MI-ANFIS networks with and without Rule Dropout.}
     \label{fig:MIANFISoverfit}
\end{figure}

In order to demonstrate the gain in generalization acquired by MI-ANFIS when utilizing Rule Dropout, we train an MI-ANFIS architecture for binary classification with and without Rule Dropout on a multiple instance dataset sampled from COREL \cite{andrews2002support}. The dataset classify whether an image contains an elephant or not, and consists of 200 images (bags): 100 positive images containing the target animal and 100 negative images containing other animals. Each image is represented as a set of patches (instances) and each patch is in turn represented by 230 features describing color, texture and shape information. Before training, we applied PCA to reduce the dimensionality of the features to 10 dimensions to speedup MI-ANFIS. Table \ref{dataDropout} summarizes the dataset characteristics.  Two MI-ANFIS networks composed of 15 rules each, with one network employing Rule Dropout (with $p=0.7$, this hyper-parameter was selected based on trial and error), were trained on 90\% of the data, and the remaining 10\% was used for testing (split was done randomly). Figure \ref{fig:MIANFISoverfit} shows the training and testing errors for both networks during 100 epochs. As it can be seen, without Rule Dropout, starting at epoch 20, testing performance begins to degrade while training error continues to decrease. In other words, overfitting begins to occur. Typically, using a cross validation data set, this point can be detected and training would be stopped. However, this assumes that a cross validation data is available (or training data is large enough to be split into training and testing) and more important that the cross validation data is representative of the testing data.  On the other hand, when using Rule Dropout, overfitting is significantly reduced and MI-ANFIS achieved better testing performance at the end of the training phase. Even though, when using Rule Dropout the training and testing error rates oscillate (due to the randomness of the dropout process), overall MI-ANFIS achieved 0.1123 testing SSE with Rule Dropout compared to 0.1451 testing SSE without Rule Dropout.

\begin{table}[!t]
\renewcommand{\arraystretch}{1.1}
\caption{Benchmark data sets}
\label{dataDropout}
\centering
\begin{tabular}{l||c|c|c|c|c}
\hline
\bfseries Data set & \bfseries dim.(PCA) & \bfseries No. Bags & \bfseries Positive & \bfseries Negative &\bfseries No.Instances\\
\hline\hline
Elephant & 230(10)& 200& 100& 100&  $2 \rightarrow 13$\\
\hline
\end{tabular}
\end{table}

\subsection{Application To Landmine Detection}
\begin{figure}[h!]
   \centerline{
        \includegraphics[width=0.6\linewidth]{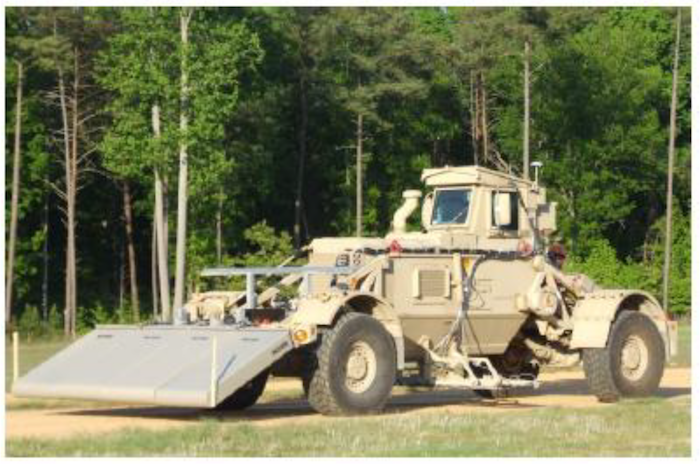}
    }
    \caption{Vehicle mounted GPR system for detecting buried Landmines.}
    \label{fig:GPRve}
\end{figure}
\begin{figure}[htb]
\centering
\subfigure[3D GPR Raw data.]{

  \includegraphics[width=0.2\linewidth]{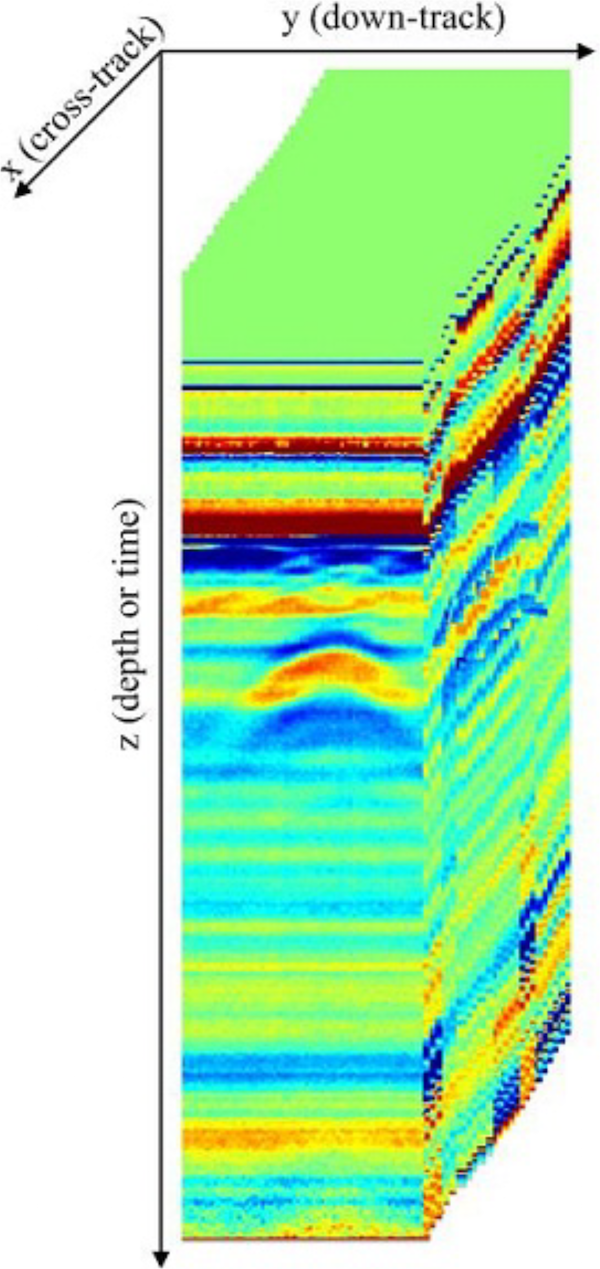}

  \label{3D}
}
\subfigure[2-D views of the raw GPR data.]{
  \includegraphics[width=0.3\linewidth]{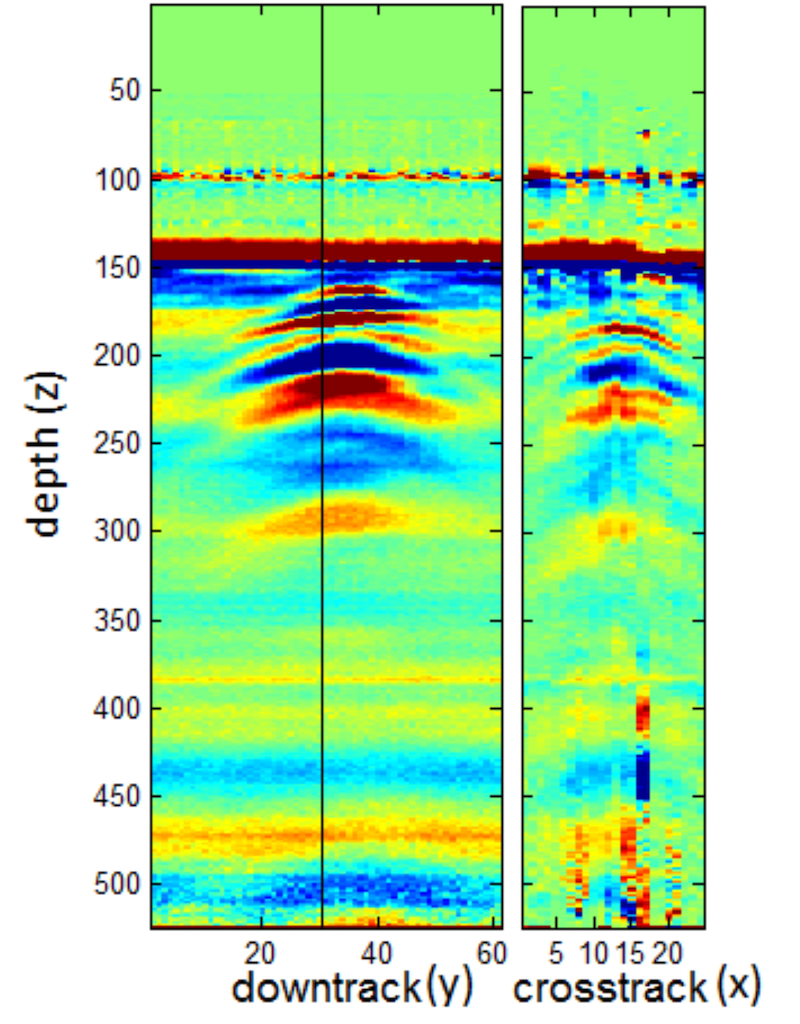}

  \label{2D}
}
\caption{3-dimensional and 2-dimensional raw GPR data.}
\label{GPRdata}
\end{figure}

In this section, we report the results of applying the proposed Multiple Instance Inference to fuse the output of multiple discrimination algorithms for the purpose of landmine detection using Ground Penetrating Radar (GPR). GPR data collected at different locations and different dates were used to train and test the proposed MI-ANFIS. The alarm collection covers 319 encounters of various anti-tank mines with high metal content (ATHM) and 422 encounters of various anti-tank mines with low metal content (ATLM).
The vehicle-mounted GPR sensor collects 3-dimensional data as the vehicle moves (Figure \ref{fig:GPRve}). The 3-dimensions correspond to the spatial location on the ground (down-track, cross-track, and depth) and is shown in Figure \ref{GPRdata}.
Figure \ref{2D} shows a 2-D views of (depth, down-track) and (depth, cross-track) slices of GPR data. As it can be seen, the target signature does not extend over all depth values. Thus, one global feature vector may not discriminate between mines and clutter effectively. To overcome this limitation, most classifiers developed for this application extract multiple features from small overlapping windows at multiple depths. In the following, we assume that each training alarm (3-D data cube) has been divided into 15 overlapping (depth wise) patches. Each patch is processed by 2 discrimination algorithms. These algorithms are based on the Edge Histogram Descriptor (EHD) \cite{EHDpknn}. The first one, called EHDDT, extracts features from each 2-D (down-track, depth) patch. The second discrimination algorithm, called EHDCT, extracts information for the 2-D (Cross-Track, depth) patch. In addition, auxiliary features are synthesized from each patch. In particular, \quotes{SignatureWidth} in the Down-track direction and \quotes{SignatureWidth} in the Cross-Track direction are used to capture the effective width of the hyperbolic shape within each patch. These auxiliary features are intended to provide contextual information that can support the relevance of the EHDDT and/or EHDCT. As a result, each alarm is represented by a Bag of 15 instances and each instance is a 4-dimensional feature vector. Each bag is labeled as positive (has a target) or negative (non target), but labels at the instance level are not available. The X-Y Ground truth information about the target is available (using GPS and known target position on calibration lanes). However, the depth position cannot be easily identified as it depends on target size, burial depth, soil type, and other environmental conditions. Manually extracting the depth location can be very tedious. Similarly, during testing, it is not trivial how to combine partial confidence values from the multiple windows. Therefore, the MIL paradigm is suitable to solve this problem.
%
%
%
%

\begin{figure}[h!]
    \centerline{
        \includegraphics[width=0.9\linewidth]{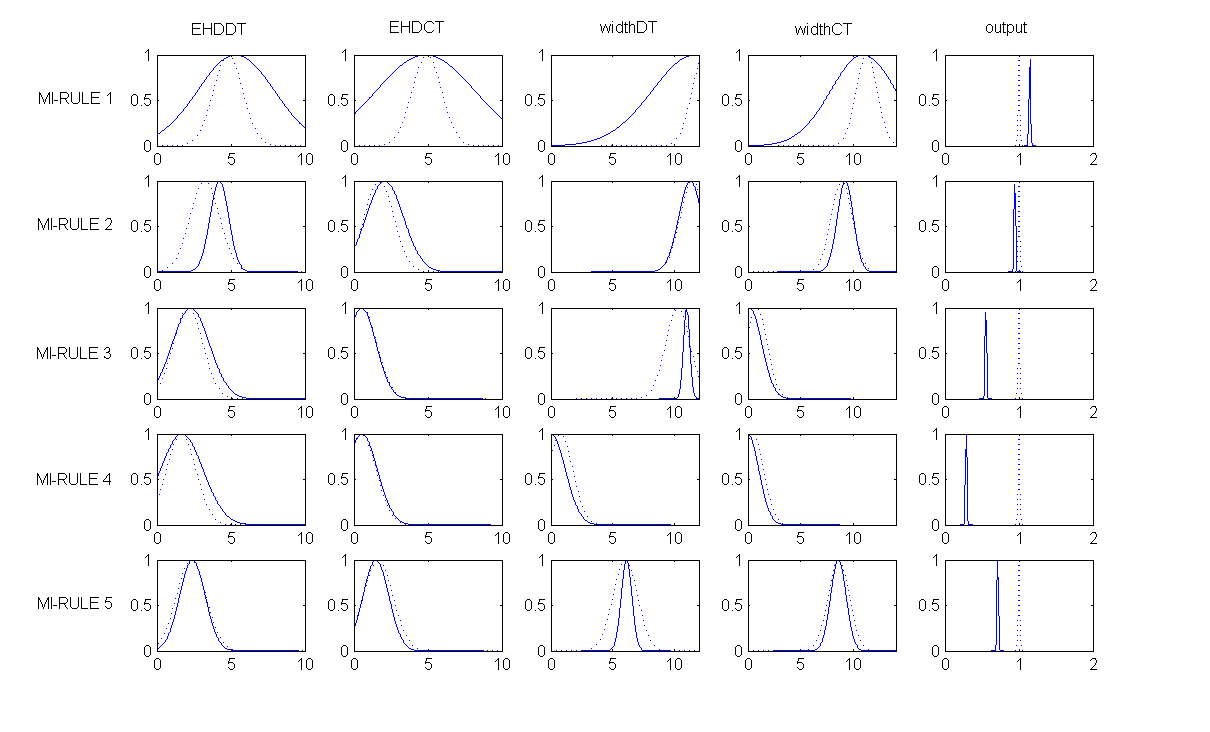}
    }
    \caption{MI-ANIFS fusion rules before (dotted lines) and after training (solid lines).}
    \label{fusionEpoch150}
\end{figure}

\normalsize
We construct a zero-order MI-ANFIS (constant consequent parameters) having 5 multiple instance rules, and employing Gaussian MFs to describe the input fuzzy sets. To initialize the system's parameters, first, we use the FCM algorithm to cluster the instances that belong to positive bags into 5 clusters, and we initialize the MFs' centers as the clusters' centers. Then, we initialize the standard deviations of the input MFs and the output parameters to 1.
\\After initialization, we run MI-ANFIS basic learning algorithm (Algorithm \ref{alg1}) to jointly learn a fuzzy description of the positive concepts as well as optimal rules' output. Figure \ref{fusionEpoch150} is a graphical representation of the 5 multiple instance rules prior to running the optimization process (dotted line curves) and the learned rules after training (continuous curves). The fuzzy sets of the rules' antecedents describe the location and the extent of the positive concepts in the 4-D instance feature space. The rules' consequent values can be interpreted as an assessment of the \quotes{positivity} of each learned concept. For instance, the MI-ANFIS learned the following two positive concepts to describe targets:

\footnotesize
\begin{equation}
\nonumber
\mathcal{R}^1: \; \mbox{If} \; EHDDT \; \mbox{is} \; Medium \; \mbox{and} \; EHDCT \; \mbox{is} \; Medium\; \mbox{and} \; WidthDT \; \mbox{is} \; High \;
 \mbox{and} \; WidthCT \; \mbox{is} \; High \; \mbox{then} \; o^{1}=1.15.
\label{eq:rule1fusion}
\end{equation}
\begin{equation}
\nonumber
\mathcal{R}^2: \; \mbox{If} \; EHDDT \; \mbox{is} \; Medium \; \mbox{and} \; EHDCT \; \mbox{is} \; Low\;   \mbox{and} \; WidthDT \; \mbox{is} \; High \;
 \mbox{and} \; WidthCT \; \mbox{is} \; High \; \mbox{then} \; o^{2}=0.94.
\label{eq:rule1fusion}
\end{equation}

\normalsize

%
%
%
%

\begin{figure}[h!]
    \centerline{
        \includegraphics[width=0.8\linewidth]{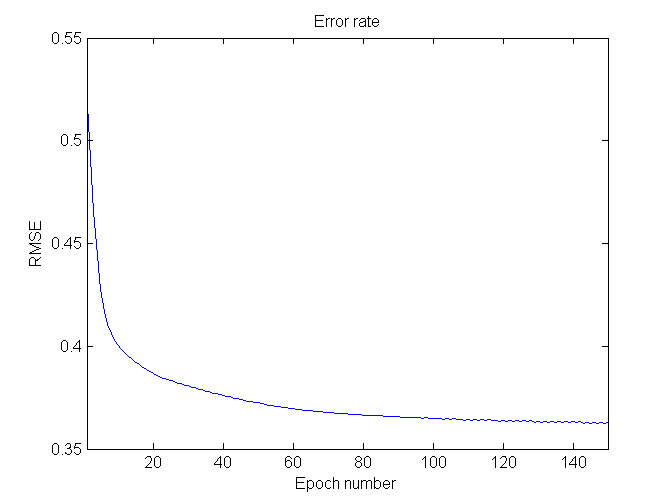}
    }
    \caption{A plot of MI-ANFIS RMSE during 150 training epochs.}
    \label{fig:ErrorrateMIANFIS}
\end{figure}

\subsubsection{Results}
\begin{figure}[h!]
  \centering
  \includegraphics[width=0.9\linewidth]{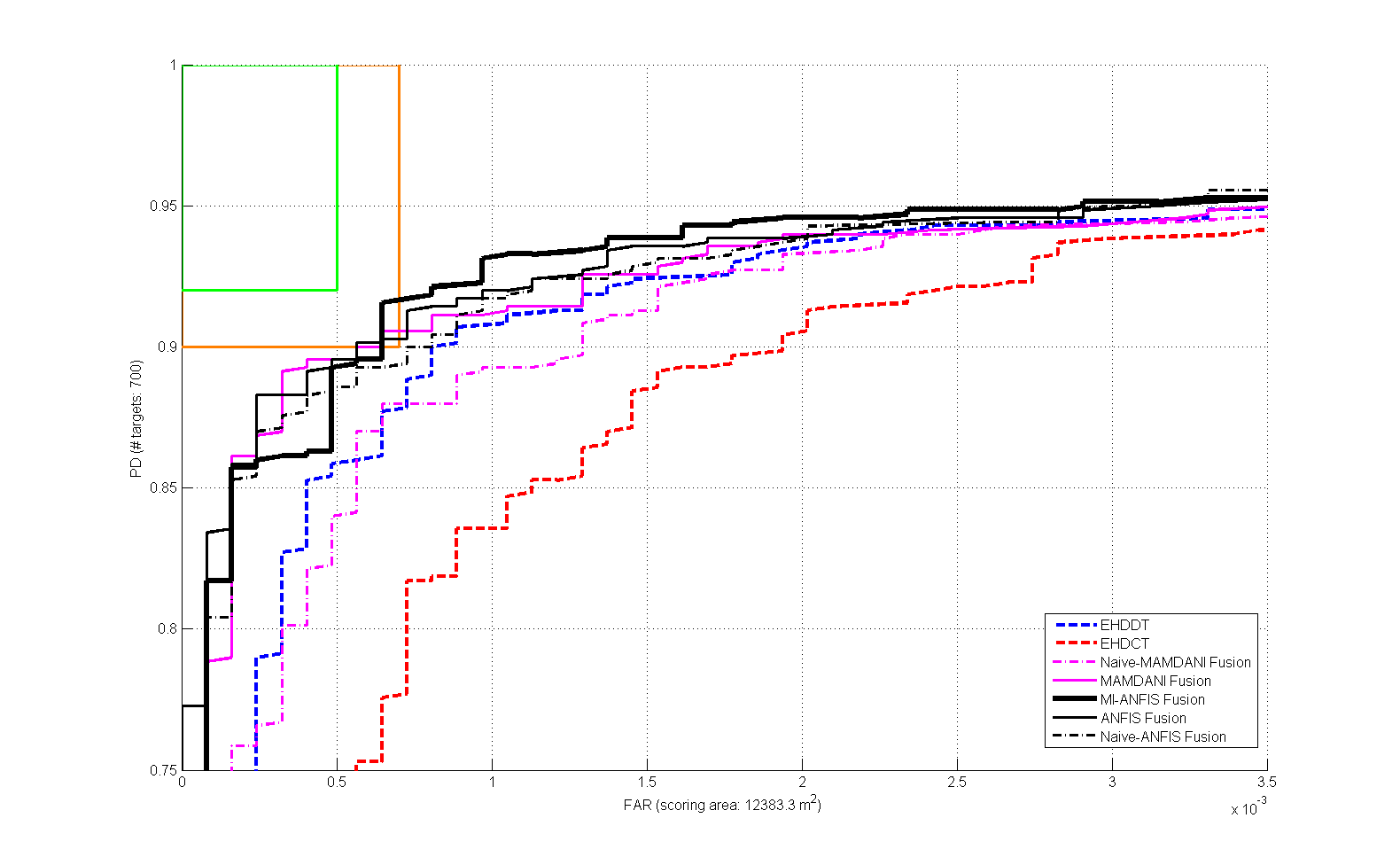}
   \caption{Comparison of the individual discriminators, the proposed MI-ANFIS, Mamdani, ANFIS, NaiveMamdani, and NaiveANFIS fusion methods. Note that the Mamdani and ANFIS systems have the advantage of instance-level labels on training data.}
     \label{fig:ROC}
\end{figure}

The proposed fusion method was trained and tested using 10-folds cross validation. Figure \ref{fig:ROC} displays the ROCs (averaged over the 10 folds).
To provide a quantitative evaluation of the proposed multiple instance fuzzy inference fusion method, we compare its performance to a fusion method based on the standard Mamdani \cite{KhalifaSPIE2014} and standard ANFIS \cite{khalifa2014fusion}. Since the standard Mamdani and AFNIS cannot learn from partially labeled data, an expert is used to label all instances of all bags within the training data. We also compare MI-ANFIS performances to a naive MIL implementation of Mamdani (NaiveMamdani) and ANFIS (NaiveANFIS) where all instances from positive bags are considered positive and all instances from negative bags are considered negative.\\
As it can be seen in Figure \ref{fig:ROC}, MI-ANFIS performed better than the standard ANFIS on the large dataset, and as expected NaiveANFIS performed worse. The standard ANFIS performed better at low FAR (False Alarms Rate), the reason is that strong Mines are easy to identify manually and in this case, the  ground truth helps. However, weaker mine signatures are not as easy to localize, so the truth may not be as accurate and can degrade the performance. Overall, MI-ANFIS outperformed all presented fusion approaches and the individual discriminators (EHDDT and EHDCT). This is due to the ability of MI-ANFIS to overcome labeling ambiguity by learning meaningful concepts.

As in standard ANFIS, we cannot prove convergence of the algorithms. However, in all conducted experiments MI-ANFIS converged in less than 150 epochs. Figure \ref{fig:ErrorrateMIANFIS} plots the root mean squared error (RMSE) vs. the training epoch number.

\section{Related Work}\label{Related-work}
MI-ANFIS deals with ambiguity by introducing the novel concept of \textit{truth instances}: when carrying reasoning using a bag of instances at Layer 2 (Figure \ref{fig:MI-Anfis-arch}), a proposition will not only have one degree of truth, it will have multiple degrees of truth ($r_{ij}$), we call truth instances. Thus, effectively encoding the third vagueness component of ambiguity and increasing the expressive power of traditional fuzzy logic. In addition to effectively model ambiguity, MI-ANIFS has the inherited capability of assessing the prediction quality by outputting soft values. For example, depending on the $\alpha$ parameter of Softmax in Layer 3, MI-ANFIS can assign higher outputs to bags with more than one positive instance. Thus, giving the end user a way to assess the positiveness of a given bag. 
\\Learning positive target concepts from ambiguously labeled data has been the core task of various MIL algorithms (e.g. Diverse Density \cite{maron1998learning}). MI-ANFIS has proven that it can learn positive concepts effectively while jointly providing a fuzzy representation of such regions. The fuzzy representation is combined into meaningful and simple multiple instance rules that can be easily visualized and interpreted.
\\Compared to previously proposed multiple instance neural networks, such as Multiple Instance Neural Networks \cite{zhou2002neural} (MI-NN) and Multiple Instance RBF Neural Networks \cite{zhang2006adapting} (RBF-MIP), MI-ANFIS advantage is the use of multiple instance fuzzy logic to learn a fuzzy representation of true positive concepts. MI-NN only learns standard neural network weights that do not carry any information regarding target concepts. On the other hand, while standard RBF neural networks have been shown to be equivalent to zero order traditional Sugeno systems under certain constraints \cite{JangRBF}, thus, capable of learning a fuzzy representation of the inputs, RBF-MIP networks have different architecture and they do not employ adaptive radial basis functions in the first layer. Instead, they represent the inputs by computing their distances to clusters of training bags. This latter method is computationally expensive and its success depends greatly on the quality of the training data as it takes into consideration all the training examples which may include wrongly (noisily) labeled bags. RBF-MIP learns only discriminative regions of the bags space and does not learn true positive concepts. Moreover, MI-ANFIS learning algorithms can be updated to support a wide range of loss functions (criterions) such as cross entropy \cite{yi2004automated}, maximum margin \cite{li2006efficient}, etc. MI-NN is designed to use a handcrafted loss function which is largely responsible for the multiple instance behavior of the system and cannot be changed without substantially changing the architecture of MI-NN. This could be disadvantageous if MI-NN is to be used to solve multiple instance - multiple class classification problems.

\section{Conclusions}
In this paper, we have introduced a new framework to accomplish fuzzy inference with multiple instance data. Our work generalizes Sugeno fuzzy inference style to reason with multiple instances, the new inference style is called MI-Sugeno. We then used MI-Sugeno to develop MI-ANFIS, a novel neuro-fuzzy architecture that extends the standard Adaptive Neuro-Fuzzy Inference System (ANFIS) to reason with bags of instances in order to solve multiple instance learning problems. We developed a BackPropagation learning algorithm and showed that the proposed system is capable of learning meaningful concepts from ambiguously labeled data.
\\MI-ANFIS deals with ambiguity by introducing the novel concept of truth instances: when carrying reasoning using a bag of instances at Layer 2 (Figure \ref{fig:MI-Anfis-arch}), a proposition will not only have one degree of truth, it will have multiple degrees of truth ($r_{ij}$), we call truth instances. Thus, effectively encoding the third vagueness component of ambiguity and increasing the expressive power of traditional fuzzy logic.
\\Learning  positive concepts from ambiguously labeled data has been the core task of various MIL algorithms. MI-ANFIS has proven that it can learn positive concepts effectively while jointly providing a fuzzy representation of such regions. The fuzzy representation is combined into meaningful and simple multiple instance rules that can be easily visualized and interpreted.
\\Using synthetic and benchmark data sets we showed that the proposed Multiple Instance Fuzzy Inference is comparable to state of the art MI machine learning algorithms. We also used our framework for a real application and applied it to fuse the output of multiple discrimination algorithms for the purpose of landmine detection using Ground Penetrating Radar.
 \\In situations where overfitting is imminent, for example when using relatively smaller datasets to learn very large MI-ANFIS networks, we proposed a regularization technique, we called Rule Dropout, and showed that it could be used to train MI-ANFIS systems with better generalization.
 \\In future work, we intend to develop a multiple class version of MI-ANFIS to be used to solve multiple class classification problems. In addition, we will conduct a detailed analysis of MI-ANFIS convergence.


%

\appendices
\section{derivation of premise parameters update rules}\label{updateeq1}
From equations (\ref{eq:bpepsilonthetac}) and (\ref{eq:bpepsilonthetasigma}) the error rate for the premise parameters $c_{kj}$ and $\sigma_{kj}$ are defined as following
\begin{equation}
\frac{\partial E_p}{\partial c_{kj}}=\sum_{i=1}^{M_p}\frac{\partial E_p}{\partial O_{(1,i+[(k-1)D+(j-1)]M_p)}}\frac{\partial O_{(1,i+[(k-1)D+(j-1)]M_p)}}{\partial c_{kj}}. \nonumber
\label{eq:bpepsilonthetacbis}
\end{equation}
and,
\begin{equation}
\frac{\partial E_p}{\partial \sigma_{kj}}=\sum_{i=1}^{M_p}\frac{\partial E_p}{\partial O_{(1,i+[(k-1)D+(j-1)]M_p)}}\frac{\partial O_{(1,i+[(k-1)D+(j-1)]M_p)}}{\partial \sigma_{kj}}. \nonumber
\label{eq:bpepsilonthetasigmabis}
\end{equation}
Using the chain rule defined in (\ref{eq:bpepsilonothernodes}), we have
\begin{multline}
\frac{\partial E_p}{\partial O_{(1,i+[(k-1)D+(j-1)]M_p)}}=\frac{\partial E_p}{\partial O_{(6,1)}}\times\frac{\partial O_{(6,1)}}{\partial O_{(5,k)}}\times\frac{\partial O_{(5,k)}}{\partial O_{(4,k)}}\times\frac{\partial O_{(4,k)}}{\partial O_{(3,k)}}\times\frac{\partial O_{(3,k)}}{\partial O_{(2,i+(k-1)M_p)}}\times\frac{\partial O_{(2,i+(k-1)M_p)}}{\partial O_{(1,i+[(k-1)D+(j-1)]M_p)}}.
\label{eq:bpepsilonthetacin}
\end{multline}
Hence, (\ref{eq:bpepsilonthetac}) is equivalent to
\begin{multline}
\frac{\partial E_p}{\partial c_{kj}}=\frac{\partial E_p}{\partial O_{(6,1)}}\times\frac{\partial O_{(6,1)}}{\partial O_{(5,k)}}\times\frac{\partial O_{(5,k)}}{\partial O_{(4,k)}}\times\frac{\partial O_{(4,k)}}{\partial O_{(3,k)}}\\
\times\sum_{i=1}^{M_p}\Bigg[\frac{\partial O_{(3,k)}}{\partial O_{(2,i+(k-1)M_p)}}\times\frac{\partial O_{(2,i+(k-1)M_p)}}{\partial O_{(1,i+[(k-1)D+(j-1)]M_p)}}\times\frac{\partial O_{(1,i+[(k-1)D+(j-1)]M_p)}}{\partial c_{kj}}\Bigg].
\label{eq:bpepsilonthetacf}
\end{multline}
From (\ref{eq:bpepsilon6}), we have
\begin{equation}
\frac{\partial E_p}{\partial O_{(6,1)}}=-2(t_p-O_p).
\label{eq:bpepsilon6bis}
\end{equation}
It is also straightforward to show that,
\begin{equation}
\frac{\partial O_{(6,1)}}{\partial O_{(5,k)}}=\frac{\partial (\sum_{i=1}^{|O_3|} O_{(5,i)})}{\partial O_{(5,k)}}=1.
\label{eq:bpepsilon65}
\end{equation}
and,
\begin{multline}
\frac{\partial O_{(5,k)}}{\partial O_{(4,k)}}=\frac{\partial (\overline{w}_k\mathcal{S}_{\alpha}(\mathbf{x}_{p1}\cdot \mathbf{b}^k,\;  \mathbf{x}_{p2}\cdot \mathbf{b}^k, \ldots, \mathbf{x}_{pM_p}\cdot \mathbf{b}^k))}{\partial (\overline{w}_k)}=\mathcal{S}_{\alpha}(\mathbf{x}_{p1}\cdot \mathbf{b}^k,\;  \mathbf{x}_{p2}\cdot \mathbf{b}^k, \ldots, \mathbf{x}_{pM_p}\cdot \mathbf{b}^k).
\label{eq:bpepsilon54}
\end{multline}
Continuing with the derivation, we have
\begin{equation}
\frac{\partial O_{(4,k)}}{\partial O_{(3,k)}}=\frac{\partial \overline{w}_k}{\partial w_k}= \frac{\partial \Big(\frac{w_i}{\sum_{l=1}^{|O_3|}w_l}\Big)}{\partial w_k}= \frac{ \sum_{l=1}^{|O_3|}w_l - w_k}{\Big(\sum_{l=1}^{|O_3|}w_l\Big)^2}.
\label{eq:bpepsilon43}
\end{equation}
and,
\begin{multline}
\frac{\partial O_{(3,k)}}{\partial O_{(2,i+(k-1)M_p)}}=\frac{\partial \mathcal{S}_{\alpha}(\{r_{(k,j)}\}_{j=1}^{M_p})}{\partial r_{(k,(i+(k-1)M_p))}}=\frac{e^{\alpha r_{(k,(i+(k-1)M_p))}}}{\sum_{m=1}^{M_p} e^{\alpha r_{(k,m)}}}\Big[1+\alpha\Big(r_{(k,(i+(k-1)M_p))}-\mathcal{S}_{\alpha}(\{r_{(k,m)}\}_{m=1}^{M_p})\Big)\Big].
\label{eq:bpepsilon32}
\end{multline}
The details of the derivation of the \quotes{softmax} function details can be found at \cite{Maron}.
\\Next, we need to compute$ \frac{\partial O_{(2,i+(k-1)M_p)}}{\partial O_{(1,i+[(k-1)D+(j-1)]M_p)}}$. We have,
\begin{equation}
O_{(2,i+(k-1)M_p)}=\prod_{d=1}^D\mu_{A_{\Big(\ceil[\big]{(i+(k-1)M_p)/ M_p},d\Big)}}\Big(x_{(p,(i+(k-1)M_p)[M_p],d)}\Big).
\label{eq:bpepsilon211}
\end{equation}
and,
\begin{equation}
O_{(1,i+[(k-1)D+(j-1)]M_p)}=\mu_{A_{(k,j)}}(x_{(p,(i+(k-1)M_p)[M_p],d)}).
\label{eq:bpepsilon212}
\end{equation}
Thus,
\begin{multline}
\frac{\partial O_{(2,i+(k-1)M_p)}}{\partial O_{(1,i+[(k-1)D+(j-1)]M_p)}} =\\\frac{\partial \Bigg(\prod_{d=1}^D\mu_{A_{\Big(\ceil[\big]{(i+(k-1)M_p)/ M_p},d\Big)}}\Big(x_{(p,(i+(k-1)M_p)[M_p],d)}\Big)\Bigg)}{\partial \Big(\mu_{A_{(k,j)}}(x_{(p,(i+(k-1)M_p)[M_p],d)})\Big)}
= \prod_{d=1, d\neq j}^D\mu_{A_{\Big(\ceil[\big]{(i+(k-1)M_p)/ M_p},d\Big)}}\Big(x_{(p,(i+(k-1)M_p)[M_p],d)}\Big).
\label{eq:bpepsilon213}
\end{multline}
Finally, we have
\begin{multline}
\frac{\partial O_{(1,i+[(k-1)D+(j-1)]M_p)}}{\partial c_{kj}}  =\frac{\partial \mu_{A_{(k,j)}}(x_{(p,(i+(k-1)M_p)[M_p],j)})}{\partial c_{kj}}=\\
\frac{(x_{(p,(i+(k-1)M_p)[M_p],j)}-c_{kj})}{\sigma_{kj}^2}\times exp({-\frac{(x_{(p,(i+(k-1)M_p)[M_p],j)}-c_{kj})^2}{2\sigma_{kj}^2}}).
\label{eq:bpepsilon1c}
\end{multline}
Substituting the derivatives in (\ref{eq:bpepsilonthetacf}), we obtain (\ref{eq:bpepsilonthetacval}).

The update formula for $\sigma_{kj}$ can be derived in a similar manner. It can be shown that
\begin{equation}
\begin{split}
\frac{\partial E_p}{\partial \sigma_{kj}}& = -2(t_p-O_p)\times \mathcal{S}_{\alpha}(\mathbf{x}_{p1}\cdot \mathbf{b}^k,\;  \mathbf{x}_{p2}\cdot \mathbf{b}^k, \ldots, \mathbf{x}_{pM_p}\cdot \mathbf{b}^k) \times \frac{ \sum_{l=1}^{|O_3|}w_l - w_k}{\Big(\sum_{l=1}^{|O_3|}w_l\Big)^2}\\
& \times \sum_{i=1}^{M_p} \Bigg(\frac{e^{\alpha r_{(k,(i+(k-1)M_p))}}}{\sum_{m=1}^{M_p} e^{\alpha r_{(k,m)}}}\Big[1+\alpha\Big(r_{(k,(i+(k-1)M_p)})-\mathcal{S}_{\alpha}(\{r_{(k,m)}\}_{m=1}^{M_p})\Big)\Big] \\ & \times \prod_{d=1, d\neq j}^D\mu_{A_{\Big(\ceil[\big]{(i+(k-1)M_p)/ M_p},d\Big)}}\Big(x_{(p,(i+(k-1)M_p)[M_p],d)}\Big)
\\ & \times \frac{(x_{(p,(i+(k-1)M_p)[M_p],j)}-c_{kj})^2}{\sigma_{kj}^3}\times exp({-\frac{(x_{(p,(i+(k-1)M_p)[M_p],j)}-c_{kj})^2}{2\sigma_{kj}^2}}) \Bigg).
\label{eq:bpepsilonthetacvalapx}
\end{split}
\end{equation}

\section{derivation of consequent parameters update rule}\label{updateeq2}
The error rate for the consequent parameters is defined in equations (\ref{eq:bpepsilonthetconsBall}) and (\ref{eq:bpepsilonthetconsB}). Next, we compute $\frac{\partial E_p}{\partial O_{(5,i)}}$ using the previously defined chain rule in (\ref{eq:bpepsilonothernodes}), and obtain
\begin{equation}
\frac{\partial E_p}{\partial O_{(5,i)}}=\frac{\partial E_p}{\partial O_{(6,1)}}\times\frac{\partial O_{(6,1)}}{\partial O_{(5,i)}}.
\label{eq:bpepsilonthetconsBchain}
\end{equation}
From (\ref{eq:bpepsilon6}), we have
\begin{equation}
\frac{\partial E_p}{\partial O_{(6,1)}}=-2(t_p-O_p).
\label{eq:bpepsilon62}
\end{equation}
And from (\ref{eq:bpepsilon65}), we have
\begin{equation}
\frac{\partial O_{(6,1)}}{\partial O_{(5,i)}}=1.
\label{eq:bpepsilon65}
\end{equation}
Continuing with the derivation, we have
\begin{multline}
\frac{\partial O_{(5,i)}}{\partial b_j^i}  =\frac{\partial (\overline{w}_i\mathcal{S}_{\alpha}(\mathbf{x}_{p1}\cdot \mathbf{b}^i,\;  \mathbf{x}_{p2}\cdot \mathbf{b}^i, \ldots, \mathbf{x}_{pM_p}\cdot \mathbf{b}^i))}{\partial b_j^i}
=\frac{\partial}{\partial b_j^i} \Big(\overline{w}_i\sum_{m=1}^{M_p} \frac{\mathbf{x}_{pm}\cdot \mathbf{b}^i exp(\alpha \mathbf{x}_{pm}\cdot \mathbf{b}^i)}{\sum_{h=1}^{M_p} exp(\alpha \mathbf{x}_{ph}\cdot \mathbf{b}^i)}\Big)\\=\overline{w}_i\sum_{m=1}^{M_p} \frac{\partial}{\partial b_j^i} \Big( \frac{\mathbf{x}_{pm}\cdot \mathbf{b}^i exp(\alpha \mathbf{x}_{pm}\cdot \mathbf{b}^i)}{\sum_{h=1}^{M_p} exp(\alpha \mathbf{x}_{ph}\cdot \mathbf{b}^i)}\Big)
=\overline{w}_i\sum_{m=1}^{M_p}\frac{1}{\Big(\sum_{h=1}^{M_p} exp(\alpha(\mathbf{x}_{ph}\cdot \mathbf{b}^i-\mathbf{x}_{pm}\cdot \mathbf{b}^i))\Big)^2}\\\times\Big[\Big(x_{(p,m,j)}\sum_{h=1}^{M_p} exp(\alpha(\mathbf{x}_{ph}\cdot \mathbf{b}^i-\mathbf{x}_{pm}\cdot \mathbf{b}^i)\Big)
-\Big(\mathbf{x}_{pm}\cdot \mathbf{b}^i\sum_{h=1}^{M_p} exp(\alpha(\mathbf{x}_{ph}\cdot \mathbf{b}^i-\mathbf{x}_{pm}\cdot \mathbf{b}^i)\alpha(x_{(p,h,j)}-x_{(p,m,j)})\Big)\Big].
\label{eq:bpepsilonthetconsB5b}
\end{multline}
Thus, the overall error rate with respect to the consequent parameter $b_{j}^i$ is given according to (\ref{eq:bpepsilonthetaall}) in equation (\ref{eq:bpepsilonthetaallbij}).

\section*{Acknowledgment}
This work was supported in part by U.S. Army Research Office Grants Number W911NF-13-1-0066 and W911NF-14-1-0589. The views and conclusions contained in this document are those of the authors and should not be interpreted as representing the official policies, either expressed or implied, of the Army Research Office, or the U.S. Government.

\ifCLASSOPTIONcaptionsoff
  \newpage
\fi



%
\bibliographystyle{IEEEtran}
\bibliography{proposalBib}

\end{document}